%% file: main.tex
\documentclass[10pt,twocolumn,letterpaper]{article}

\usepackage[final,algorithms]{wacv}      % To produce the REVIEW version for the algorithms track
\usepackage{wacv}              % To produce the CAMERA-READY version
\input{preamble}

\definecolor{wacvblue}{rgb}{0.21,0.49,0.74}
\usepackage[pagebackref,breaklinks,colorlinks,allcolors=wacvblue]{hyperref}

\def\wacvPaperID{2174} % *** Enter the WACV Paper ID here
\def\confName{WACV}
\def\confYear{2026}

\title{Equivariant Sampling for Improving Diffusion Model-based Image Restoration}

\author{Chenxu Wu$^{1,2}$\qquad
Qingpeng Kong$^{1,2}$\qquad
Peiang Zhao$^{1,2}$ \qquad
Wendi Yang$^{1,2}$ \qquad
\\
Wenxin Ma$^{1,2}$ \qquad
Fenghe Tang$^{1,2}$ \qquad
Zihang Jiang$^{1,2}$\footnotemark[1] \qquad 
S.Kevin Zhou$^{1,2,3,4}$\thanks{Corresponding author.}
\\
$^1$ School of Biomedical Engineering, Division of Life Sciences and Medicine, USTC\\
$^2$ MIRACLE Center, Suzhou Institute for Advance Research, USTC \\
$^3$  Key Laboratory of Intelligent Information Processing of CAS, ICT, CAS \\
$^4$ State Key Laboratory of Precision and Intelligent Chemistry, USTC \\
{\tt\small wuchenxu@mail.ustc.edu.cn jzh0103@ustc.edu.cn s.kevin.zhou@gmail.com}
}

\begin{document}
\twocolumn[{
\renewcommand\twocolumn[1][]{#1}
\maketitle
\vspace{-12mm}
\begin{center}
    \captionsetup{type=figure}
    \includegraphics[width=1\linewidth]{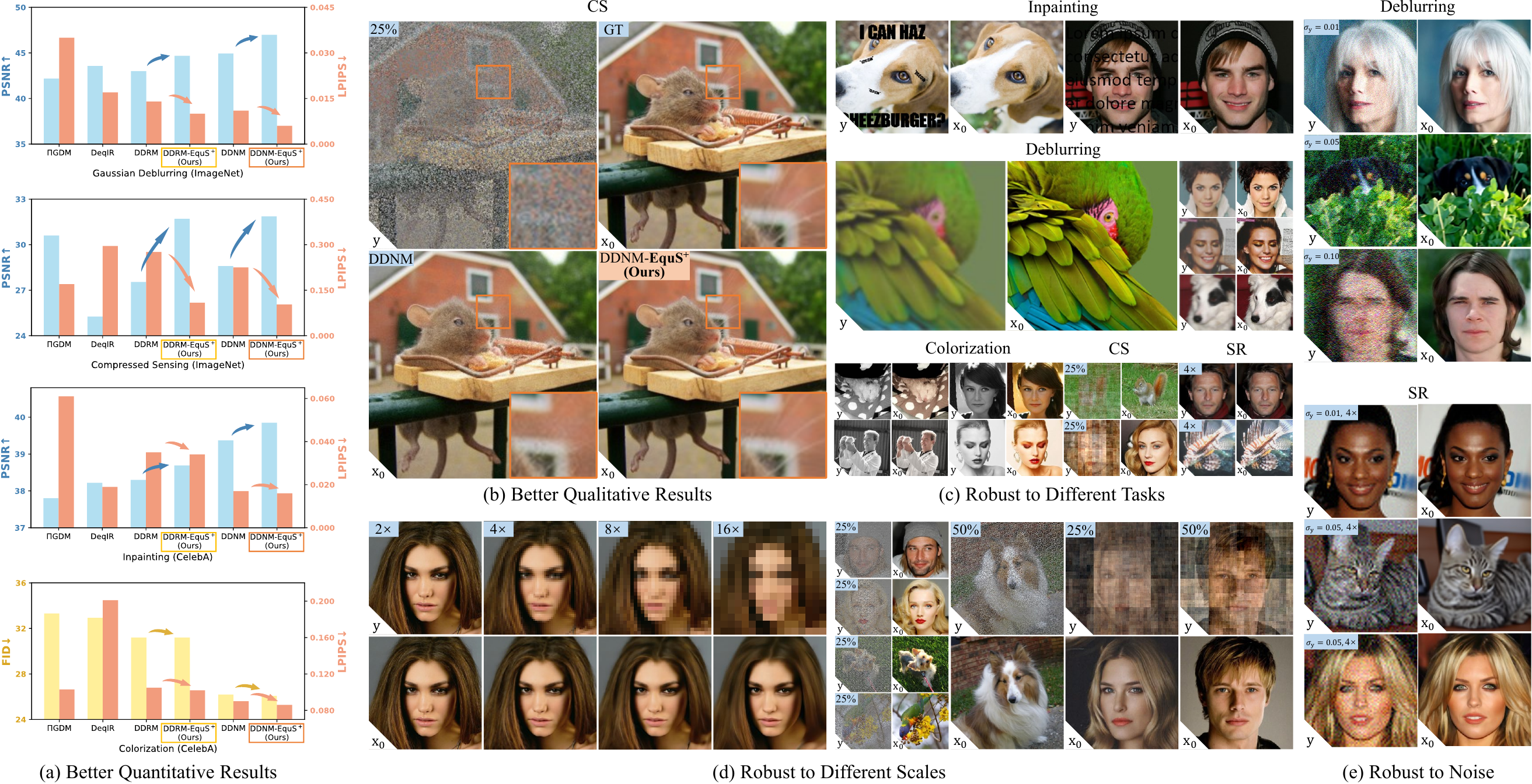}
    \vspace{-5mm}
    \captionof{figure}{Our method offers (a) superior quantitative performance, (b) improved qualitative results. It is (c) adaptable to various IR applications, (d) robust to different scales, and (e) resilient to different noise levels. $\y$ represents the degraded image, $\x_0$ denotes the sampling result, SR represents super-resolution and CS represents compressed-sensing.}
    \label{fig:overview}
\end{center}
}]
\input{sec/0_abstract}

\input{sec/1_intro}

\input{sec/2_related_work}
\input{sec/3_preliminary}
\input{sec/4_methods}

\input{sec/5_experiments}
\input{sec/6_conclusion}


{
    \normalem
    \small
    \bibliographystyle{ieeenat_fullname}
    \bibliography{main}
}

\end{document}

%% file: preamble.tex
\usepackage{algorithm,algpseudocode}
\usepackage{wrapfig}

\input{defs_mine_iccv}

\usepackage{pifont}

\def\ie{\mbox{\textit{i.e.}, }}
\def\eg{\mbox{\textit{e.g.}, }}

\usepackage{multicol}
\usepackage{multirow}
\usepackage{colortbl}
\usepackage{ulem}
\usepackage{xcolor}

%% file: defs_mine_iccv.tex
 {\everymath{\displaystyle\everymath{}}\array}%
 {\endarray}
\everymath{\displaystyle\everymath{}}

\newcommand{\ntransf}{|\mathcal{F}|}

\def\A{\mathbf{A}}

\def\F{\mathcal{F}}

\def\I{\mathbf{I}}

\def\x{\mathbf{x}}
\def\y{\mathbf{y}}

\def\bmu{\boldsymbol{\mu}}

\def\0{\mathbf{0}}
\def\1{\mathbf{1}}

\newcommand{\denoiser}{\mathcal{D}_\theta}

%% file: sec/0_abstract.tex
% \begin{figure*}
%     \centering
%     \includegraphics[width=1\linewidth]{}
%     \caption{Caption}
%     \label{fig:overveiw}
% \end{figure*}
\begin{abstract}
Recent advances in generative models, especially diffusion models, have significantly improved image restoration (IR) performance. However, existing problem-agnostic diffusion model-based image restoration (DMIR) methods face challenges in fully leveraging diffusion priors, resulting in suboptimal performance. In this paper, we address the limitations of current problem-agnostic DMIR methods by analyzing their sampling process and providing effective solutions. We introduce \textbf{EquS}, a DMIR method that imposes equivariant information through dual sampling trajectories. To further boost EquS, we propose the \textbf{T}imestep-\textbf{A}ware \textbf{S}chedule (\textbf{TAS}) and introduce \textbf{EquS$^+$}. TAS prioritizes deterministic steps to enhance certainty and sampling efficiency. Extensive experiments on benchmarks demonstrate that our method is compatible with previous problem-agnostic DMIR methods and significantly boosts their performance without increasing computational costs. Our code is available in the Supplementary.
\end{abstract}

%% file: sec/1_intro.tex
\section{Introduction}
\label{sec:intro}
%-------------------------------------------------------------------------
% \begin{figure*}[h]
%     \centering
%     \includegraphics[width=1\linewidth]{pic/main_pic.pdf}
%     \caption{Caption}
%     \label{fig:overveiw}
% \end{figure*}
%-------------------------------------------------------------------------
% \begin{figure*}
%   \centering
%   \begin{subfigure}{0.5\linewidth}
%     \includegraphics[width=1\linewidth]{pic/motivation1.pdf}
%     \caption{Trajectory of original sampling.}
%     \label{fig:geometry-a}
%   \end{subfigure}
%   \hfill
%   \begin{subfigure}{0.49\linewidth}
%     \includegraphics[width=1\linewidth]{pic/motivation2.pdf}
%     \caption{Trajectory of FTS sampling.}
%     \label{fig:geometry-b}
%   \end{subfigure}
%   \caption{Conceptual illustration of the trajectories of two different sampling processes. $\mathcal{H}$ represents the contours of the data distribution. FTS improves the final sampling results by using bidirectional score sampling, bringing them closer to $\x$.}
%   \label{fig:diagram}
% \end{figure*}

%-------------------------------------------------------------------------
\begin{figure*}
\centering
\includegraphics[width=0.99\textwidth]{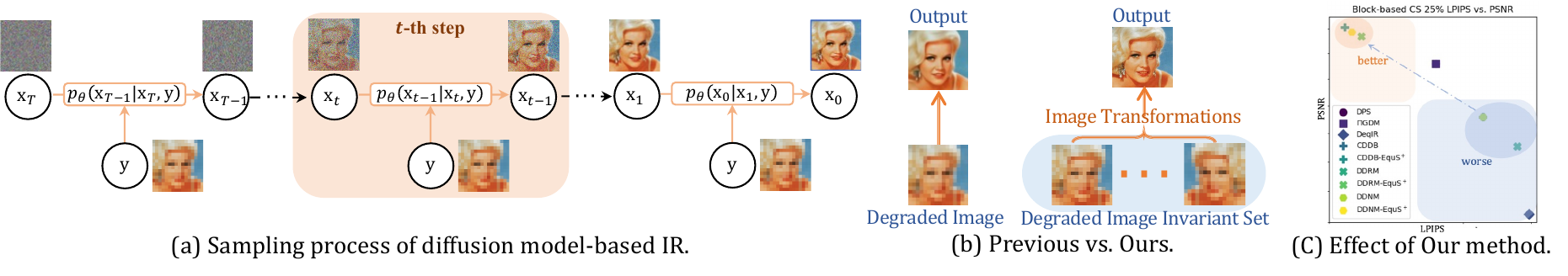}
\caption{\small
(a) Sampling process of diffusion model-based IR. (b) Previous vs. Ours. (c) Effect of our methods.}
\label{fig:previous}
\end{figure*}

Image restoration (IR) aims to recover the original image from a degraded image. 
It remains a longstanding challenge due to its ill-posed nature and significant practical applications \cite{richardson1972bayesian, banham1997digital}. 
Traditional IR methods \cite{dong2015srcnn,dong201arcnn,cao2023ciaosr,cao2022datsr,cao2022davsr,restor15,AOTGAN,lin2019dudonet} are limited to single-task solutions, requiring multiple deep neural network (DNN) trainings. Recently, diffusion models~\citep{ho2020ddpm, sohl2015deep, song2019generative, song2020score} have emerged as powerful generative models~\cite{dhariwal2021diffusion, rombach2022high}, offering a fresh approach to IR.
Their ability to model complex, high-dimensional distributions positions them as effective solutions for addressing problem-specific IR challenges \cite{croitoru2023diffusion}, encompassing applications such as super-resolution~\cite{li2022srdiff, gao2023implicit, saharia2022sr3, choi2021ilvr}, deblurring~\cite{chen2024hierarchical, ren2023multiscale}, inpainting~\cite{corneanu2024latentpaint, meng2021sdedit, lugmayr2022repaint}, colorization~\cite{carrillo2023diffusart, weng2024cad}, and compressed sensing~\cite{yang2024lossy, chung2023solving}.

%-------------------------------------------------------------------------
% 最近有人研究 agnostic 的 DMIR ，

Notably, a growing family of research  \cite{kawar2022ddrm, chung2023dps, wang2022ddnm, cao2024deep, zhu2023diffpir, alccalar2024zero, garber2024image, dou2024diffusion,lee2024diffusion} leverages diffusion models to address various IR tasks in a zero-shot manner. 
These problem-agnostic diffusion model-based image restoration (DMIR) methods can utilize diffusion priors from pre-trained diffusion models irrespective of the specific problem at hand. They achieve this by incorporating the degraded image via inverse guidance or conditioning during the sampling process, as shown in Figure \ref{fig:previous} (a).
However, these methods face challenges in fully leveraging diffusion priors, leading to suboptimal results.
% These problem-agnostic diffusion model-based image restoration (DMIR) methods can utilize generative priors irrespective of the specific problem at hand as they incorporate the degraded image via inverse guidance during the sampling process.
%However, these studies have two main issues: \textbf{(1)} they rely on unidirectional guidance in the sampling process, and \textbf{(2)} they use a uniform sampling schedule. As a result, they do not effectively leverage generative priors.
% 这些方法有关键的问题（具体的问题是什么？写明白），因为他们没有很好的使用prior

%-------------------------------------------------------------------------
% 之前的方法有什么问题 
\textbf{First}, these methods guide the sampling process based on unidirectional information, using a single sampling trajectory. The reliance on a single degraded image for determining the restoration output, as illustrated in Figure \ref{fig:previous} (b), may hinder performance by introducing biases into the reconstructed image. Furthermore, when the degraded image is noisy, the stochasticity in the noise can exacerbate errors within a single sampling trajectory. Even minor errors may propagate through the long sampling chain inherent to diffusion models, leading to artifacts and reduced image quality.
% 想一下其他的weakness 1.measurement 2.随机性太强了 单个误差
% where capturing information from multiple perspectives is essential for accurate reconstruction. 
% 因此，使用均一的方案有问题 加一点其他的问题 举例子
\textbf{Second}, previous methods implement a uniform sampling schedule, assigning equal importance to all time steps in the sampling process. However, in the context of DMIR, the degraded image is already close to the original image. Consequently, using a uniform schedule is suboptimal, as it concentrates too much on early steps that contribute little to the reconstruction. For instance, in tasks like super-resolution or deblurring, later steps often play a more critical role in refining details. Thus, employing a uniform schedule leads to inefficiencies and may reduce the accuracy of the final reconstruction.

In this paper, we delve into the sampling process of DMIR, introducing \textbf{Equ}ivariant \textbf{S}ampling (\textbf{EquS})—a DMIR method developed to fully leverage diffusion priors. Specifically, we incorporate an equivariant inverse mapping into the sampling process, introducing a novel approach that enables dual-trajectory sampling. By reinterpreting degraded images from an alternative perspective and facilitating their interaction with diffusion priors, EquS harnesses equivariant information to enhance reconstruction performance.
% 在前面 Second 的位置强调一下 uniform schedule，不要放后面，和这里衔接上 get this
To overcome the limitations of conventional uniform schedules in previous arts, we further propose the \textbf{T}imestep-\textbf{A}ware \textbf{S}chedule (\textbf{TAS}) and introduce an enhanced version of EquS, termed as \textbf{EquS$^+$}.
Unlike uniform schedules, 
% 为什么 prioritizes deterministic time steps 对 IR 有好处？写明原因，不要写 impactful 之类的套话
TAS prioritizes deterministic time steps, introducing more certain information into the sampling outcomes and reducing sampling randomness, which is beneficial in IR where diversity in sampling outcomes is less critical. We conduct comprehensive experiments, demonstrating the exceptional performance of EquS and EquS$^+$ across various IR tasks.

%% file: sec/2_related_work.tex
\section{Related Work}
\label{sec:Related Work}

\noindent\textbf{IR frameworks.}
Researchers have explored frameworks that combine techniques such as residual blocks \cite{VDSR,plug-denoiser,zhou2020dudornet}, attention mechanisms \cite{inpainting3,deepfillv2,restormer,MAT,NAFNet,liang2022vrt,cao2021vsrt,cao2023ciaosr,cao2022datsr,cao2022davsr}, GANs \cite{goodfellow2014generative,WGAN-GP,ESRGAN,Real-ESRGAN,inpainting-GAN,cao2018lccgan,cao2020improving,zhang2021bsrgan,menon2020pulse,pan2021exploiting,dgp}, and other methodologies \cite{restor12,restor14,restor15,AOTGAN,lin2019dudonet} to enhance IR performance. Self-supervised learning approaches, such as Noise2X \cite{lehtinen2018noise2noise,krull2019noise2void,batson2019noise2self,kim2021noise2score,moran2020noisier2noise,xu2020noisy,mansour2023zero,wang2023noise2info} and equivariant imaging (EI) \cite{chen2021equivariant,chen2022robust,chen2023imaging,terris2024equivariant}, have also been explored as powerful frameworks for tackling IR challenges.

\noindent\textbf{Problem-specific DMIR}
focuses on training conditional diffusion models either in the image space \cite{saharia2022sr3, whang2022deblurring, saharia2022palette, yue2022difface, liu2024residual, liu2023i2sb, zhang2024diffusion, dong2024building, wu2024id, li2024rethinking, liu2024cdformer, ye2024learning} or in the latent space \cite{rombach2022ldm, wang2023stablesr, xia2023diffir, lin2023diffbir}. In order to do this within a zero-shot framework, several methods utilize the diffusion priors, integrating the reference image during the sampling process \cite{choi2021ilvr,lugmayr2022repaint}. Overall, these methods predominantly concentrate on an individual IR task.

%-------------------------------------------------------------------------
\noindent\textbf{Problem-agnostic DMIR}
utilizes the diffusion priors of pre-trained models (DDPM \cite{ho2020ddpm}, DDIM \cite{song2020ddim}) as versatile, plug-and-play building blocks for solving various IR tasks.
In the design of various guidance methods, DDRM \cite{kawar2022ddrm} utilizes singular value decomposition (SVD) to the degradation operator, thereby tackling linear IR problems in a manner akin to SNIPS \cite{kawar2021snips}. DDNM \cite{wang2022ddnm} computes the pseudo-inverse matrix for range-null space decomposition \cite{schwab2019deep, wang2023gan}, iteratively refining the null space in the sampling process. These advancements \cite{wang2023unlimited, gandikota2024text, chung2024deep} further expand the applicability of DDNM. DeqIR \cite{cao2024deep} uses the guidance method in DDNM and presents a fixed-point solver for parallel sampling. DDPG \cite{garber2024image} offers a guidance method that demonstrates robustness to noise.
To further improve the IR performance, many methods consider computing deeper gradients for the guidance. For example, DPS \cite{chung2023dps} employs a Jensen approximation for posterior sampling. ZAPS \cite{alccalar2024zero} presents an alternative guidance methodology for DPS. {$\Pi$}GDM \cite{song2023pseudoinverse} introduces Jacobian calculation to pseudo-inverse guidance. CDDB \cite{chung2024direct} integrates the guidance method into I{$^2$}SB \cite{liu2023i2sb}, thereby enhancing its IR performance. However, all of these methods rely on a single sampling trajectory that considers only unidirectional information, which may lead to suboptimal performance.

%% file: sec/3_preliminary.tex
\section{Preliminaries}
\label{sec:Preliminaries}
%-------------------------------------------------------------------------

\noindent\textbf{Problem Statement.} IR focuses on restoring a high-quality image $\hat{\x}$ from a degraded image $\y = \mathbf{A}\x + \mathbf{n}$, where $\y \in \mathbb{R}^{m}$ represents the observed measurement, $\x \in \mathbb{R}^{n}$ denotes the original image, $\mathbf{n}$ represents additive nonlinear noise (\eg $\mathcal{N}(\mathbf{0},\sigma^2_{\y}\mathbf{I})$), and $\mathbf{A} \in \mathbb{R}^{m \times n}$ is a known measurement operator with $m \leq n$, \eg a bicubic downsampler. The restoration process can be formulated as
\begin{equation}
\begin{aligned}
    \hat{\x} = \underset{\x}{\operatorname{arg\, min}} 
    &\frac{1}{2\sigma^{2}} \|\mathbf{A}\x - \y\|_{2}^{2} + \lambda \mathcal{R}(\x), \quad \lambda, \sigma \in \mathbb{R},
    \label{eq:restoration}
\end{aligned}
\end{equation}
where the first term, $ \|\mathbf{A}\x - \y\|_{2}^{2}$, ensures data consistency, while the second term, $\lambda \mathcal{R}(\x)$, introduces regularization based on prior knowledge of natural image, such as Tikhonov regularization~\cite{alberti2021learning}.

%-------------------------------------------------------------------------
\noindent\textbf{Diffusion Models.} DDPM \cite{ho2020ddpm} could generate high-quality images through a forward process and a reverse process. In the forward process, noise with specified levels is introduced to the data, \ie
\begin{align}
    \label{eq:ddpm forward}
    q(\x_t | \x_{0}) := \mathcal{N} \left( \x_t; \sqrt{\bar\alpha_t} \x_{0}, (1-\bar\alpha_t) \I \right),
\end{align}
where $\bar\alpha_t:=\Pi_{s=1}^t \alpha_s$, $\alpha_t:=1-\beta_t$ and $\beta_t$ is a variance of noise level.
For the reverse process, the previous state $\x_{t-1}$ can be predicted with $\tilde{\bmu}_t$ and $\tilde{\sigma}_t$, which is formulated as
\begin{align}
    q(\x_{t-1} | \x_{t}, \x_{0}) := \mathcal{N} \left( \x_{t-1}; \tilde{\bmu}_t(\x_t, \x_0), {\sigma}_t^2 \I \right),
    \label{eq:reverse_dis}
\end{align}
where $\tilde{\bmu}_t(\x_t, \x_0):=\frac{\sqrt{\bar\alpha_{t-1}}\beta_t}{1-\bar\alpha_t} \x_0 + \frac{\sqrt{\alpha_t}(1-\bar\alpha_{t-1})}{1-\bar\alpha_t} \x_t$ and ${\sigma}_t^2 := \frac{1{-}\bar\alpha_{t{-}1}}{1{-}\bar\alpha_t} \beta_t$.
However, $\x_0$ is intractable but can be obtained from Gaussian denoising as implied in \eqref{eq:ddpm forward}, \ie
\begin{align}
    \label{eq:x0_est}
    {\x}_{0|t} & :=\frac{1}{\sqrt{\bar{\alpha}_{t}}} \left (  \x_t - \sqrt{1-\bar{\alpha}_t} \epsilon_\theta(\x_t;t) \right ),
\end{align}
where $\epsilon_\theta(\x_t;t)$ predicts the noise in $\x_t$. By iteratively calculating $\x_{t-1}$ using \eqref{eq:reverse_dis}, clean images $\x_{0}$ can be obtained from $\x_{T}$$\sim$$\mathcal{N}(\mathbf{0},\mathbf{I})$. 
% DDIM \cite{song2020ddim} replace the step in \eqref{eq:reverse_dis} with
% \begin{equation}
%     \label{eq:ddim_step}
%     \x_{t-1} := \sqrt{\bar{\alpha}_{t-1}} {\x}_{0|t} + \hat{\sigma}_t \epsilon_\theta (\x_t;t) + {\tilde{\sigma}_t \epsilon_t},
% \end{equation}
% where $\hat{\sigma}_t = \sqrt{1-\bar{\alpha}_{t-1}-\tilde{\sigma}_t^2}$. The hyperparameter $\tilde{\sigma}_t$ regulates the stochasticity of the whole reverse process.
%-------------------------------------------------------------------------
DDPM can also be described by stochastic differential equations (SDE). 
In VP-SDE \cite{song2020score}, the score-based diffusion models \cite{song2019generative,song2020score} can be viewed as a continuous-time generalization of DDPM. They satisfy the following equation
\begin{align}
    \nabla_{\x_t} \log p_t(\x_t) \simeq \mathbf{s}_\theta(\x_{t}, t) \simeq -\frac{{\epsilon}_\theta(\x_{t},{t})}{\sqrt{1-\bar{\alpha}_t}},
    \label{eq:score=ddpm}
\end{align}
where score function $\mathbf{s}_\theta(\x_{t}, t)$ models $\nabla_\x \log p_t(\x)$ with score matching methods \cite{hyvarinen2005estimation,song2019generative}. The details of this section are provided in the supplementary.

%-------------------------------------------------------------------------

\noindent\textbf{Diffusion Model-based Image Restoration.} In generation tasks, one only needs to train $\nabla_{\x_t} \log p_t(\x_t)$ (\ie to train $\epsilon_\theta(\x_{t}, t)$) to execute the reverse process. Assuming the availability of problem-specific scores for all noise levels, \ie $\nabla_{\x_t} \log p_t(\x_t|\y)$, diffusion models can address IR tasks via \eqref{eq:reverse_dis}. For problem-specific DMIR, we can directly train $\nabla_{\x_t} \log p_t(\x_t|\y)$ (\ie train $\epsilon_\theta(\x_{t}, \y, t)$). However, for problem-agnostic DMIR, the problem-specific scores can be decomposed using Bayes' theorem~\cite{chung2023dps,song2023pseudoinverse}
\begin{equation}
\nabla_{\x_t} \log p_t(\x_t|\y) = \nabla_{\x_t} \log p_t(\x_t) + \nabla_{\x_t} \log p_t(\y|\x_t),
\label{eq:score decompose}
\end{equation}
where the first term can be approximated by the score network $\mathbf{s}_\theta(\x_{t}, t)$ using \eqref{eq:score=ddpm} \cite{vincent2011connection}, and the second term acts as a guidance component, denoting the score of $p_t(\y|\x_t)$.

Although $\nabla_{\x_t} \log p_t(\y|\x_t)$ is intractable in problem-agnostic DMIR, this time-dependent likelihood can be approximated through various methodologies. We summarize these approaches in Section \ref{Elucidating the Design Space of Guidance}.

%-------------------------------------------------------------------------

%% file: sec/4_methods.tex
\section{Elucidating the Design Space of Guidance}
\label{Elucidating the Design Space of Guidance}
%-------------------------------------------------------------------------
In this section, we unify previous problem-agnostic DMIR methods within a common framework, demonstrating that, despite being motivated by different theories, the resulting algorithms show minor differences. We include detailed explanation of this section in the supplementary.
%-------------------------------------------------------------------------

In the case of VP-SDE or DDPM sampling,  
$p_\theta(\x_0|\x_t)$ has the unique posterior mean
\begin{align}
\label{eq:post}
 \x_{0|t} := \mathbb{E}[\x_0|\x_t] &= \frac{1}{\sqrt{{\bar\alpha_t}}}(\x_t + (1 - {\bar\alpha_t})\nabla_{\x_t} \log p_t(\x_t)).
 \vspace{-3pt}
\end{align}
By plugging \eqref{eq:score=ddpm} into \eqref{eq:post}, one can retrieve the empirical Bayes optimal posterior mean \cite{kim2021noise2score,chung2023dps,song2020score} $\x_{0|t}$ from $p_\theta(\x_0|\x_t)$ using either $\mathbf{s}_\theta(\x_{t}, t)$ or ${\epsilon}_\theta(\x_{t},{t})$.

% %-------------------------------------------------------------------------
% \begin{proof}
%     All proofs in this section are in Supplementary.
% \end{proof}

%-------------------------------------------------------------------------

In DMIR, one can obtain the posterior mean $\tilde\x_{0|t}$ by replacing $\nabla_{\x_t} \log p(\x_t)$ with $\nabla_{\x_t} \log p(\x_t|\y)$ in \eqref{eq:post}, \ie
\begin{align}
\label{eq:post_probelm_specific}
\tilde\x_{0|t} := \mathbb{E}[\x_0|\x_t,y]= \frac{1}{\sqrt{{\bar\alpha_t}}}(\x_t + (1 - {\bar\alpha_t})\nabla_{\x_t} \log p_t(\x_t|\y)).
\end{align}
% Plugging \eqref{eq:score decompose} into \eqref{eq:post_probelm_specific} also allows us to obtain $\tilde\x_{0|t}$ from $p_\theta(\x_0|\x_t,\y)$.
%-------------------------------------------------------------------------
%Problem-agnostic DMIR necessitates the design of various solutions to approximate $\nabla_{\x_t} \log p_t(\x_t|\y)$. 
% We utilize different notations for these guidance mechanisms, denoted as $\mathcal{G}$\footnote{To simplify, we ignore scaling constants related to measurement noise.}, while ignoring the predefined terms preceding them. After which \eqref{eq:ddim_step} can be replaced with
% \begin{align}
%     \x_{t-1} = \sqrt{\bar{\alpha}_{t-1}} 
%     \underbrace{({\x}_{0|t}+ \mathcal{G})}_{{\hat\x}_{0|t}: \ \rm corrected \ {\x}_{0|t}} + \underbrace{\hat{\sigma}_t \epsilon_\theta (\x_t;t)}_{\rm noise \ in \ \x_t} + \underbrace{{\tilde{\sigma}_t \epsilon_t}}_{\rm random \ noise}.
%     \label{eq:ddim_step_with_guidance}
% \end{align}
% To address problem-agnostic DMIR with $\mathcal{G}$, previous works generally adhere to the procedure outlined in Algorithm.

\begin{figure}
  \centering
    \includegraphics[width=1\linewidth]{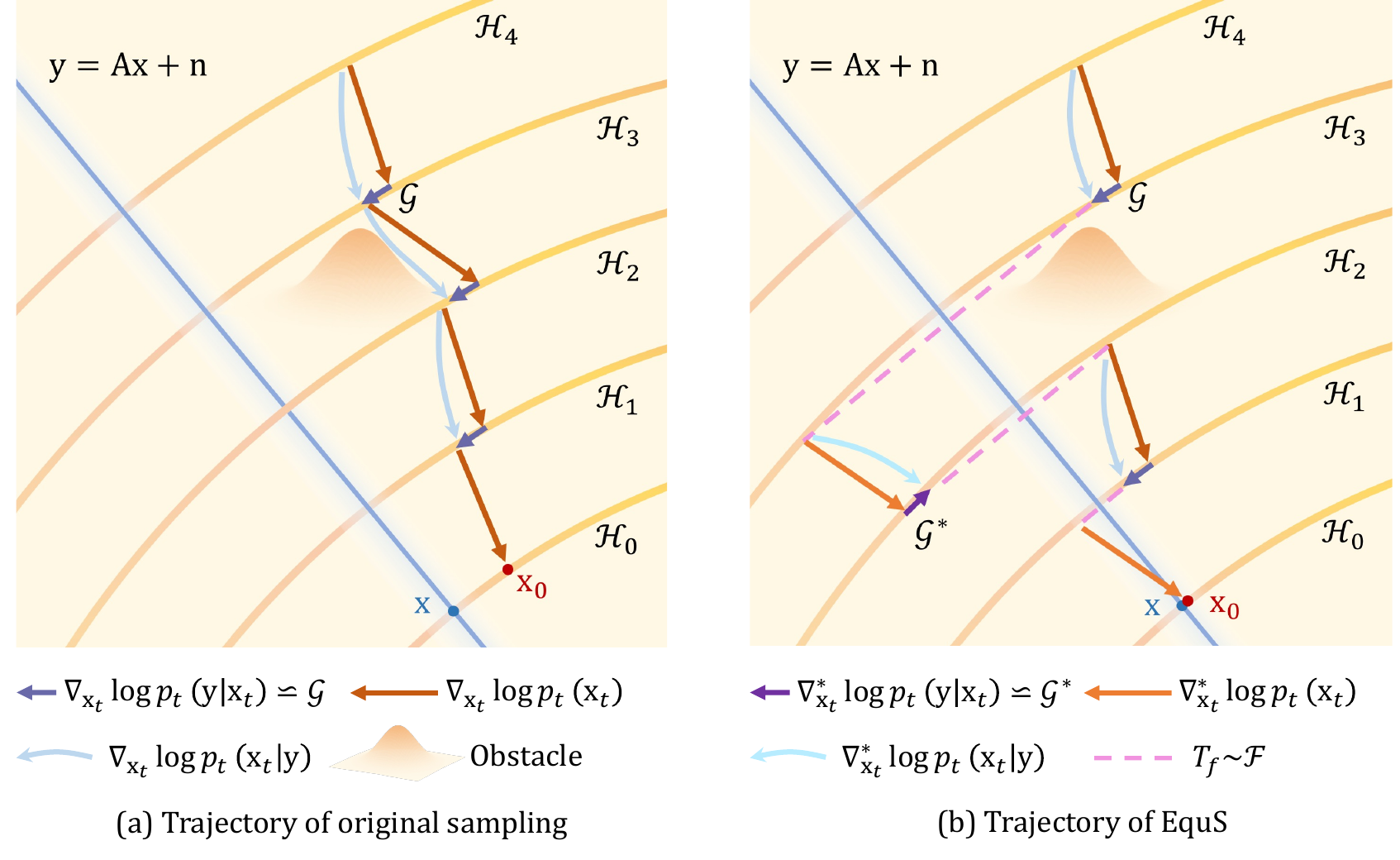}
  \caption{\small
  Conceptual illustration of the trajectories of two different sampling processes. $\mathcal{H}$ represents the contours of the data distribution. EquS enables dual-trajectory sampling, determining $\x_0$ by considering bidirectional information.}
  \label{fig:geometry}
\end{figure}
%-------------------------------------------------------------------------
However, we only have $\nabla_{\x_t} \log p_t(\x_t)$ in problem-agnostic DMIR, thus to approximate $\nabla_{\x_t} \log p_t(\y|\x_t)$ in \eqref{eq:score decompose}, various methods can be broadly classified into
\begin{align}
    \label{eq:ddnm_approx}
    \nabla_{\x_t} \log p_t(\y|\x_t)&\simeq \A^{\dagger}(\y - \A\x_{0|t}),\\
    \label{eq:cddb_approx}
    &\simeq \A^{\top}(\y - \A\x_{0|t}).
    % \label{eq:dps_approx}
    % \nabla_{\x_t} \log p_t(\y|\x_t)\simeq g\mathcal{G}_2 &= \sqrt{\bar{\alpha}_t}\underbrace{\frac{\partial \x_{0|t}}{\partial \x_t}}_{\text{\rm Jacobian}} \underbrace{\A^{\top}(\y - \A\x_{0|t})}_{\text{\rm vector}},\\
    % \label{eq:pigdm_approx}
    % \nabla_{\x_t} \log p_t(\y|\x_t)\simeq g\mathcal{G}_1 &= \sqrt{\bar{\alpha}_t}\underbrace{\frac{\partial \x_{0|t}}{\partial \x_t}}_{\text{\rm Jacobian}} \underbrace{\A^{\dagger}(\y - \A\x_{0|t})}_{\text{\rm vector}},
\end{align}
where $\A^{\dagger}$ denotes the pseudo-inverse of $\A$, $\A^{\top}$ represents the transpose of $\A$.
%These methods all share a common characteristic, \ie the term $(\y - \A\x_{0|t})$ ensures the data consistency mentioned in \eqref{eq:restoration}.
% %-------------------------------------------------------------------------
% \begin{proof}
%     Please refer to the proofs in Supplementary.
% \end{proof}
%-------------------------------------------------------------------------
 % As these approaches leverage gradient backpropagation, we denote them with a $g$ prefix before $\mathcal{G}$.

Problem-agnostic DMIR can be considered\footnote{We disregard the predefined terms preceding $\nabla_{\x_t} \log p_t(\y|\x_t)$.} as first obtaining $\x_{0|t}$ based on \eqref{eq:post}, and then correcting it by adding guidance (\eqref{eq:ddnm_approx} or \eqref{eq:cddb_approx}), \ie
\begin{align}
\tilde\x_{0|t} =  \underbrace{\frac{1}{\sqrt{{\bar\alpha_t}}}(\x_t + (1 - {\bar\alpha_t})\nabla_{\x_t} \log p_t(\x_t))}_{\rm original \ \x_{0|t} \ in \ \eqref{eq:post}} + \underbrace{\nabla_{\x_t} \log p_t(\y|\x_t)}_{\rm guidance}.
    \label{eq:post_with_guidance}
\end{align}
In Figure \ref{fig:geometry} (a), we show a conceptual illustration of the trajectory of this sampling process. To further promote IR performance, these methods \cite{chung2023dps,alccalar2024zero,song2023pseudoinverse,chung2024direct} consider computing Jacobian-vector products for \eqref{eq:ddnm_approx} and \eqref{eq:cddb_approx}, which can be implemented by backpropagation as $\nabla_{\x_t} \|\y - \A\x_{0|t}\|_2^2$ \cite{chung2023dps}. Unfortunately, while they sometimes improve the performance of IR, they also bring a higher computational burden due to the need for backpropagation.

%-------------------------------------------------------------------------
%-------------------------------------------------------------------------
%The Jacobian can be seen as taking a deeper guidance beyond $\mathcal{G}_1$ and $\mathcal{G}_2$. We summarize these methods in Table \ref{table: 4 guidance}.
\section{Equivariant Sampling}
\label{Equivariant Sampler}
Images are often assumed to possess certain invariance properties concerning specific transformation groups, including rotations, translations, and reflections. Intuitively, if an algorithm can reconstruct an image from a degraded image, it should similarly reconstruct the image from a transformed version of that degraded image. In this section, we illustrate how to integrate this concept into DMIR.
% In this section, we analyze the issues of existing problem-agnostic DMIR methods based on $\mathcal{G}_2$\footnote{The other cases can be deduced from this.}.
% In this section, we reframe the DMIR sampling within the context of the EI framework~\cite{chen2021equivariant, chen2022robust, chen2023imaging, zhao2024equivariant} and propose \textbf{Equ}ivariant \textbf{S}ampling (\textbf{EquS}).
% EquS enables dual-trajectory sampling, integrating bidirectional information. Our theoretical analysis shows that EquS allows sampling beyond the limited space of a single sampling trajectory.
% Recently, the EI framework~\cite{chen2021equivariant,chen2022robust,chen2023imaging,zhao2024equivariant} is proved to be an efficient self-supervised learning method. 
%-------------------------------------------------------------------------
%Our observations indicate that prior approaches focus on developing various methods to approximate $\nabla_{\x_t} \log p_t(\x_t|\y)$. In this section, we are considering a mechanism that is compatible with both $\nabla_{\x_t} \log p_t(\x_t)$ and $\nabla_{\x_t} \log p_t(\x_t|\y)$. 
%\footnote{Note that the range space could also be defined as $\range=\text{range}(\A^{\dagger})$}

We denote the inverse mapping in \eqref{eq:post} as $\x_{0|t} = \denoiser (\x_t)$. The guidance in Section \ref{Elucidating the Design Space of Guidance} ensures data consistency in the measurement domain
\begin{equation} 
\label{eq:data consistency}
    \y - \A\x_{0|t} = \y - \A \denoiser (\x_t),
\end{equation}
which allows us to solve the problem in \eqref{eq:restoration}.
%Since the dimension of $\A$ is lower than that of the original image ($m \leq n$), the measurement operator $\A \in \mathbb{R}^{m \times n}$ has a non-trivial linear nullspace $\nullA \subseteq \mathbb{R}^{n}$ with a dimension of at least $n - m$. For all $v \in \nullA$, we have $\A v = 0$. The complement of $\nullA$ is the range space $\range = \text{range}(\A^{\top})$, such that $\range \oplus \nullA = \mathbb{R}^{n}$, which ensures that for all $v \in \range$, we have $\A v \neq 0$ \cite{chen2021equivariant}.
%-------------------------------------------------------------------------
% However, guidance cannot capture information about $\x_t$ outside the range space of the operator $\A^{\top}$. This is because any function $v(\x_t)$ produced by $\denoiser (\x_t)$ that belongs to the nullspace of $\A$ will have $\A v(\x_t) = 0$. Only the information in range space $\range$ will be reserved.
% The measurement consistency can be expressed as $\A \denoiser(\x_t) = \A\A^{\top}\x_t + \A v(\x_t)$ where the first term is $\x_t$ as $\A\A^{\top}$ is the identity matrix, and $\A v(\x_t)=0$ for any $v(\x_t)$ in $\nullA$. Thus, $\mathcal{G}_2$ cannot capture information about $\x_t$ outside the range space of the operator $\A^{\top}$.
%-------------------------------------------------------------------------
%As shown in Section, there are multiple functions $f_\theta$ that can verify \eqref{eq:data consistency}, even if we have infinitely many samples $y_i$.

%-------------------------------------------------------------------------
\noindent\textbf{Invariant Set.}
For a group of transformations $\F = \{ f_1, \dots, f_{\ntransf} \}$ which is composed of unitary matrices $T_f \in \mathbb{R}^{n \times n}$, $\mathcal{X}$ is an invariant set when $T_f x \in \mathcal{X}$ holds for $\forall x \in \mathcal{X}$ and $\forall f \in \mathcal{F}$ \cite{chen2021equivariant,chen2022robust,chen2023imaging}.
%We assume that the DDPM sampling trajectory $\mathcal{X} = \{\x_t\}_0^T$ and the degraded image remains invariant under particular
%\footnote{While it is feasible to define transformations using non-unitary matrices \cite{serre1977linear}, we concentrate on unitary matrices (case of rotations, translations, and reflections).}
%-------------------------------------------------------------------------
\begin{figure*}
    \centering
    \includegraphics[width=1\linewidth]{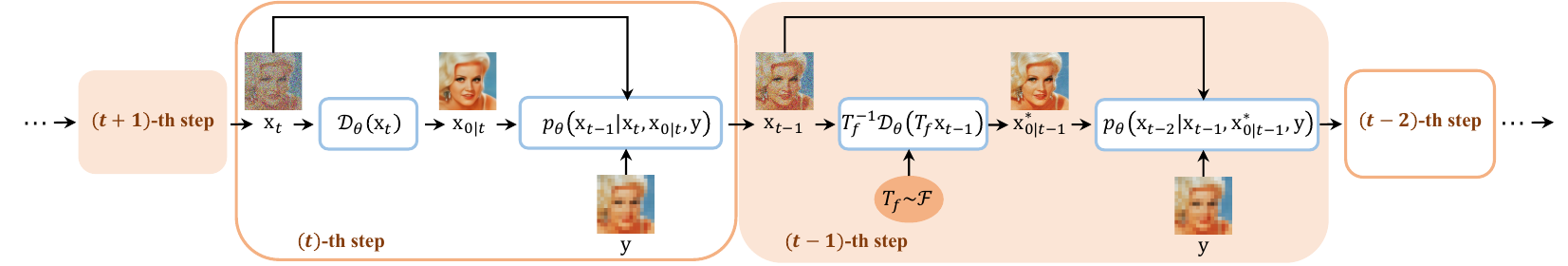}
    \caption{\small
    The sampling process of the \textbf{Equ}ivariant \textbf{S}ampling (EquS).}
    \label{fig:EquS}
\end{figure*}
%-------------------------------------------------------------------------
%-------------------------------------------------------------------------

\noindent\textbf{Equivariant Inverse Mapping.}
We say that $\denoiser$ is equivariant to the transformation group $\mathcal{F}$ if
\begin{equation}\label{eq:equivariant set consistency}
    \denoiser( T_f \x_t) = T_f \denoiser( \x_t),
\end{equation}
for all $\x_t$ and $f\in \F$. Using this property, we can get the equivariant inverse mapping, \ie
\begin{equation}
\label{eq:modified x_0}
    {\x^*_{0|t}} := T^{-1}_f\denoiser( T_f \x_t),
\end{equation}
where $T^{-1}_f$ represents the inverse transformation of $T_f$. $\nabla^*_{\x_t} \log p_t(\y|\x_t)$, $\nabla^*_{\x_t} \log p_t(\x_t)$, and $\nabla^*_{\x_t} \log p_t(\x_t|\y)$ can also be computed using \eqref{eq:ddnm_approx} or \eqref{eq:cddb_approx}, \eqref{eq:score=ddpm} and \eqref{eq:score decompose}, respectively. We leave the details in the supplementary.

\noindent\textbf{Equivariant Sampling.}
A straightforward method is to average the equivariant inverse mapping in \eqref{eq:modified x_0} over the transformation group $\mathcal{F}$. However, for large groups and in light of the lengthy sampling steps in diffusion models, this approach introduces significant computational overhead, diminishing algorithmic practicality. To mitigate this, we propose \textbf{Equ}ivariant \textbf{S}ampling (\textbf{EquS}) as an efficient estimate of the averaged equivariant inverse mapping.
For simplification, we use $p_\theta(\x_{t-1} | \x_{t}, \x_{0|t}, \y)$ to represent the composition of \eqref{eq:post_with_guidance} and \eqref{eq:reverse_dis}.
EquS first performs a standard reverse process (first $\denoiser( \x_t)$, then $p_\theta(\x_{t-1} | \x_{t}, \x_{0|t}, \y)$), followed by an equivariant reverse process (first $ T^{-1}_f\denoiser( T_f \x_t)$, then $p_\theta(\x_{t-1} |  \x_{t}, \x^*_{0|t}, \y)$). This whole process is illustrated in Figure \ref{fig:EquS}. It is worth noting that EquS requires no additional NFEs or hyperparameters.

\noindent\textbf{Equivariant Sampling from a Loss Perspective.}
With the equivariant inverse mapping, we could rewrite \eqref{eq:data consistency} as
\begin{align} 
\label{eq:data consistency generalize}
 \y - \A T^{-1}_f \denoiser( T_f\x_t) = \y - \A_f \denoiser( \tilde\x_t),
\end{align}
where $\A_f = \A\circ T^{-1}_f$\footnote{$\circ$ denotes the composition of two functions} and ${\tilde\x_t}=T_f\x_t$.

We can see that EquS introduces a new loss term $\|\y - \A_f \denoiser( \tilde\x_t)\|_2^2$ for the IR problem in \eqref{eq:restoration}, provided that \( \A \) itself is not invariant under all \( T_f^{-1} \). During the diffusion reverse iteration, EquS minimizes not only \( \|\y - \A \denoiser(\x_t)\|_2^2 \) but also \( \|\y - \A_f \denoiser( \tilde\x_t)\|_2^2 \). This addition enhances the accuracy of the final reconstructed output by imposing a new constraint, which requires that the ultimate output maintains equivariance under the transformation group during the algorithmic iteration process.

We show a conceptual illustration of the trajectory of EquS in Figure \ref{fig:geometry} (b). EquS enables dual-trajectory sampling (trajectory of the standard reverse process and trajectory of the equivariant reverse process), determining the reconstructed image $\x_0$ by considering equivariant information.

\subsection{Sampling Scheme}
\label{Free Lunch in the Sampling Schedule Design}
%Diffusion models employ well-defined fixed noise schedules ($\beta_{1:T}$), such as linear or exponential ones. Recently, many advanced techniques have emerged that traverse these schedules and sample at irregular time steps \cite{dhariwal2021diffusion,karras2022elucidating} for unconditional image generation.

\begin{figure}
    \centering
    \includegraphics[width=1\linewidth]{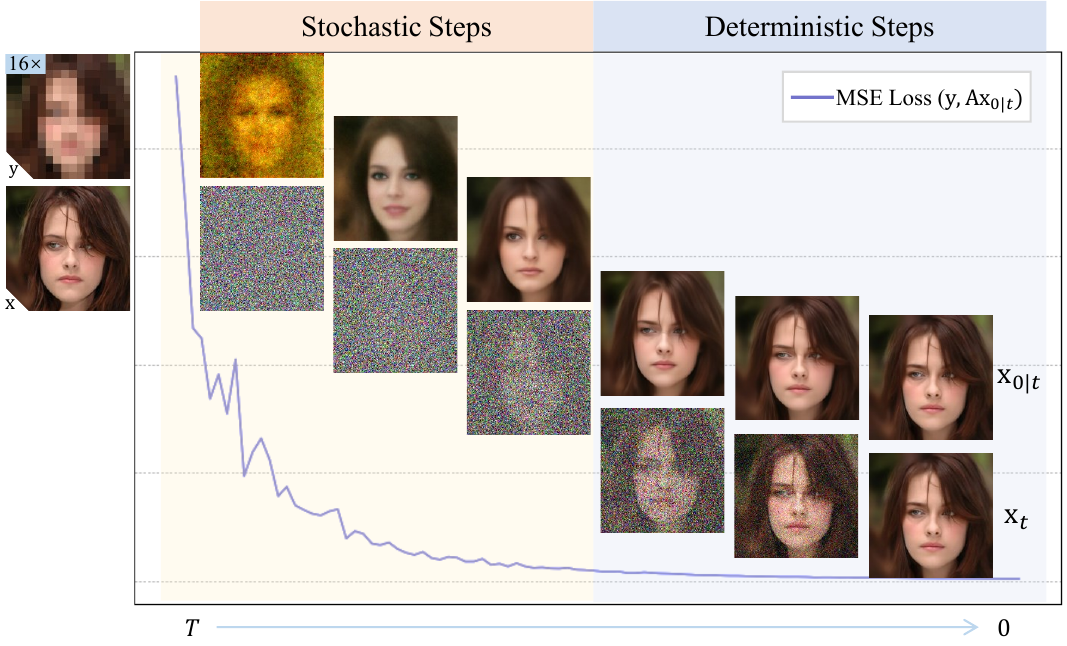}
    \caption{Deconstruction of the diffusion reverse process: stochastic steps and deterministic steps. $\x_{0|t}$ is more certain in the deterministic steps.}
    \label{fig:sampling process}
\end{figure}

\noindent\textbf{Sampling Process Deconstruction.}
As shown in Figure \ref{fig:sampling process}, the mean square error between $\A\x_{0|t}$ and $\y$ in stochastic steps is larger than that in the deterministic steps. This suggests that $\x_{0|t}$ is more certain in the deterministic steps.
Another interesting view is that we can interpret the sampling process in DMIR as a structured sequence, \ie initially mitigating residuals, followed by noise reduction \cite{liu2024residual}.

%-------------------------------------------------------------------------

%-------------------------------------------------------------------------

\noindent\textbf{Uniform Schedule Sampling.}
Most DMIR methods \cite{wang2022ddnm,kawar2022ddrm,song2023pseudoinverse,garber2024image} consider DDIM \cite{song2020ddim} to accelerate the sampling. The whole sampling could be written as the form in \eqref{eq:restoration}, \ie
\begin{equation}
\begin{aligned}
    \hat{\x} = \underset{\x}{\operatorname{arg\, min}} \sum_{t\in S}  \|\y - \A \x_{0|t}\|_2^2,
    \label{eq:restoration_ddim}
\end{aligned}
\end{equation}
where $\|\y - \A \x_{0|t}\|_2^2$ comes from the guidance in Section \ref{Elucidating the Design Space of Guidance} \cite{garber2024image}, $S$ is a sub-sequence of $[1, \ldots, T]$. However, constructing $S$ using a uniform schedule results in equal weights for the stochastic and deterministic steps. Since the term $\|\y - \A \x_{0|t}\|_2^2$ in the stochastic steps is large, this causes the summation term in \eqref{eq:restoration_ddim} to become excessively large, thus increasing the difficulty of reconstruction.

\noindent\textbf{Timestep-Aware Schedule} Based on the previous paragraph, giving more weight to the deterministic steps can reduce the summation term in \eqref{eq:restoration_ddim}, thereby introducing more deterministic information and improving the overall efficiency of the sampling process\footnote{During implementation, this can be achieved by constructing $S$, \eg using a quadratic schedule.}. Intuitively, this approach allows for finer detail refinement when $\x_{0|t}$ is deterministic. We refer to this kind of schedule, which gives more weight to the deterministic steps, as \textbf{T}imestep-\textbf{A}ware \textbf{S}chedule (\textbf{TAS}), and the version of EquS integrated with TAS is termed as EquS$^+$.
%-------------------------------------------------------------------------
%-------------------------------------------------------------------------
% \begin{figure}
%     \centering
%     \includegraphics[width=1\linewidth]{pic/loss_schedule.pdf}
%     \caption{The MSE loss plot of $\x_{0|t}$ with respect to $\x$.}
%     \label{fig:loss schedule}
% \end{figure}
%-------------------------------------------------------------------------
% \begin{figure}
%     \centering
%     \includegraphics[width=0.6\linewidth]{pic/beta_alpha.pdf}
%     \caption{Plot of different parameters varying with $t$.}
%     \label{fig:bate_alpha}
% \end{figure}

% \noindent\textbf{Observation 2.}
% The normal distribution in \eqref{eq:reverse_dis} can be expressed as
% \begin{align}
%     \x_{t-1}= \underbrace{\frac{\sqrt{\bar\alpha_{t-1}}\beta_t}{1-\bar\alpha_t} \x_{0|t}}_{\rm update} + \underbrace{\frac{\sqrt{\alpha_t}(1-\bar\alpha_{t-1})}{1-\bar\alpha_t} \x_t}_{\rm scaled \ \x_t} + \underbrace{{\sigma}_t \epsilon}_{\rm nosie}.
%     \label{eq:updata}
% \end{align}
% We observe that the reverse process from $\x_{t}$ to $\x_{t-1}$ is primarily driven by the update term in \eqref{eq:updata}, whereas the parameter preceding $\x_{0|t}$ exhibits more significant variations when $t$ is small (In Figure \ref{fig:bate_alpha}, it is evident that the slope of the parameter is considerably steep). Thus, slowing down the sampling speed at smaller $t$ values can help to balance the variations to some extent.

%% file: sec/5_experiments.tex
\begin{table*}[t]
    \centering
    % \footnotesize
    \scriptsize
    \begin{tabular}{cccccc}
        \hline
           \multicolumn{1}{c}{\rule{0pt}{10pt}\textbf{\scriptsize{ImageNet}}}&\multicolumn{1}{c}{Gaussian Deblurring} &\multicolumn{1}{c}{Block-based CS 25\%}&\multicolumn{1}{c}{Inpainting}&\multicolumn{1}{c}{{4$\times$ Bicubic SR}}&\multicolumn{1}{c}{Colorization}\\
           \rule{0pt}{10pt}Method& PSNR↑/SSIM↑/LPIPS↓ & PSNR↑/SSIM↑/LPIPS↓ &  PSNR↑/SSIM↑/LPIPS↓  &  PSNR↑/SSIM↑/LPIPS↓ & \textit{Cons}↓/FID↓/LPIPS↓\\
        \hline
            \rule{0pt}{10pt}{$\mathbf{A}^{\dagger}\mathbf{y}$} &18.57 / 0.586 / 0.376 &12.83 / 0.192 / 0.724 
            &14.62 / 0.661 / 0.413&26.27 / 0.771 / 0.302&53.92 / 44.95 / 0.167\\
            \rule{0pt}{10pt}{DPS \cite{chung2023dps}} & 26.12 / 0.729 / 0.341 & 30.61 / 0.868 / 0.170 &35.56 / 0.964 / 0.052 & 22.49 / 0.584 / 0.468&619.3 / 54.78 / 0.277 \\
            % \arrayrulecolor{blue!30}\hdashline
            % \rowcolor{blue!5}
            \rule{0pt}{10pt}{\({\Pi}\)GDM \cite{song2023pseudoinverse}}  & 42.19 / 0.980 / 0.035 & 30.61 / 0.867 / 0.170 & 35.54 / 0.964 / 0.052 & {27.89} / {0.815} / {0.217} & 177.3 / {36.75} / 0.194\\
            % \arrayrulecolor{blue!30}\hdashline
            % \rule{0pt}{10pt}{DeqIR} &43.57 / 0.986 / 0.017&21.66 / 0.510 / 0.452&31.84 / 0.954 / 0.039&26.90 /  0.788 / 0.250& 25.26 / 39.05 / 0.176\\
            \rule{0pt}{10pt}{DeqIR \cite{cao2024deep}} &43.57 / {0.986} / 0.017&25.26 / 0.696 / 0.296&32.36 / 0.956 /  0.036&26.90 /  0.788 / 0.250& {25.26} / 39.05 / 0.176\\
            % \arrayrulecolor{blue!30}\hdashline
            % \rowcolor{blue!5}
            \rule{0pt}{10pt}{CDDB \cite{chung2024direct}}  & 26.42 / 0.750 / 0.311 & 28.60 / 0.795 / 0.225& 34.92 / {0.968} / 0.026 &23.47 / 0.630 / 0.423 & 275.1 / 39.04 / 0.183\\
            \rowcolor{orange!8}
            \rule{0pt}{10pt}{CDDB\textbf{{-EquS$^+$}}}  & 27.55 / 0.797 / 0.268 & {32.05} / {0.890} / {0.099} & {36.18} / {0.976} / {0.020} & 25.41 / 0.712 / 0.358& 114.0 / 41.18 / 0.168\\
            \rule{0pt}{10pt}{DDRM \cite{kawar2022ddrm}} &43.01 / {0.986} / 0.014&27.54 / 0.766 / 0.276&34.60 / 0.965 / 0.030&27.39 / 0.801 / {0.241}&144.4 / {36.77} / {0.162}\\
            \rowcolor{orange!8}
            \rule{0pt}{10pt}{DDRM\textbf{{-EquS$^+$}}} &{44.69} / {0.990} / {0.010} &{31.70} / {0.886 } / {0.109}&{35.93} / {0.973} / {0.025}&{27.54} / {0.807} / 0.248&142.9 / 39.70 / {0.162}\\
            % \rowcolor{blue!10}
            \rule{0pt}{10pt}{DDNM \cite{wang2022ddnm}} &{44.94} / {0.990} / {0.011}&28.59 / 0.795 / 0.225  &34.93 / {0.968} / {0.025}&27.45 / 0.801 / {0.231} & {21.92} / {36.87} / {0.157}\\
            % \rowcolor{blue!5}
            % \rule{0pt}{10pt}{DeqIR-v2} & {43.60} / {0.986} / {0.017}&{21.67} / {0.510} / {0.452}
            % & {31.86} / {0.954} / {0.039} &{26.92} / {0.789} / 0.251&{25.22} / {38.96} / {0.176}\\
        
            \rowcolor{orange!8}
            \rule{0pt}{10pt}{DDNM\textbf{{-EquS$^+$}}} &{46.99 / 0.993 / 0.006} & {31.87} / {0.887} / {0.103} &{36.13} / {0.976} / {0.021}&{27.63} / {0.809} / 0.242&{21.68} / 39.34 / {0.158}\\
        \arrayrulecolor{black}\hline
    \end{tabular}
    \begin{tabular}{cccccc}
        \hline
           \multicolumn{1}{c}{\rule{0pt}{10pt}\textbf{\scriptsize{CelebA}}}&\multicolumn{1}{c}{Gaussian Deblurring} &\multicolumn{1}{c}{Block-based CS 25\%}&\multicolumn{1}{c}{Inpainting}&\multicolumn{1}{c}{{4$\times$ Bicubic SR}}&\multicolumn{1}{c}{Colorization}\\
           \rule{0pt}{10pt}Method& PSNR↑/SSIM↑/LPIPS↓ & PSNR↑/SSIM↑/LPIPS↓ & PSNR↑/SSIM↑/LPIPS↓ &  PSNR↑/SSIM↑/LPIPS↓ &  \textit{Cons}↓/FID↓/LPIPS↓  \\
        \hline
            \rule{0pt}{10pt}{$\mathbf{A}^{\dagger}\mathbf{y}$} &18.85 / 0.686 / 0.305 &12.11 / 0.155 / 0.790
            &14.65 / 0.653 / 0.459&30.64 / 0.881 / 0.214& {54.86} / 46.65 / 0.138\\
            \rule{0pt}{10pt}{DPS \cite{chung2023dps}} & 30.65 / 0.856 / 0.173& 34.02 / 0.919 / 0.122 & 37.80 / 0.967 / 0.061& 26.83 / 0.778 / 0.229 & 578.0 / 42.33 / 0.160\\
            % \rowcolor{blue!5}
            \rule{0pt}{10pt}{\({\Pi}\)GDM \cite{song2023pseudoinverse}} & 40.19 / 0.975 / 0.051 &34.02 / 0.919 / 0.122 & 37.80 / 0.966 / 0.061& 31.80 / {0.890} / {0.124}& 353.5 / 33.32 / 0.103\\
            % \rowcolor{blue!5}
            \rule{0pt}{10pt}{DeqIR \cite{cao2024deep}}&{46.85} / {0.992} / {0.011}&32.56 / 0.894 / 0.120&38.22 / {0.979} / {0.019}& {31.82} / {0.892} / 0.125&72.16 / 32.93 / 0.201\\
            \rule{0pt}{10pt}{CDDB \cite{chung2024direct}}  & 30.86 / 0.862 / 0.161 & 33.51 / 0.907 / {0.111} & 39.36 / {0.982} / {0.017} & 27.20 / 0.789 / 0.219 & 347.8 / 32.25 / 0.123\\
            \rowcolor{orange!8}
            \rule{0pt}{10pt}{CDDB\textbf{-EquS$^+$}}  & 32.60 / 0.898 / 0.162 & {34.78} / {0.927} / {0.091} & {39.82} / {0.983} / {0.016} & 30.16 / 0.855 / 0.209 & 235.4 / 32.43 / 0.110\\

            % \rowcolor{blue!10}
            \rule{0pt}{10pt}{DDRM \cite{kawar2022ddrm}} &43.07 / 0.985 / {0.023} &32.50 / 0.896 / 0.128&38.30 / 0.973 / 0.035&31.63 / 0.886 / {0.123}&170.6 / {31.20} / 0.105\\
            \rowcolor{orange!8}
            \rule{0pt}{10pt}{DDRM\textbf{-EquS$^+$}} &43.15 / {0.986} / {0.023} &{34.47} / {0.925} / {0.103}& 38.69 / 0.975 / 0.034&{32.39} / {0.901} / 0.134&169.4 / {31.20} /{0.102}\\
            % \rowcolor{blue!5}

            % \rowcolor{blue!5}
            % \rule{0pt}{10pt}{DeqIR-v2} &{46.90} / {0.993 } / {0.011}&30.33 / 0.850 / 0.174
            % &37.98 / {0.979} / {0.020}&{31.95} / {0.894} / {0.124}&71.58 / 32.82 / 0.200\\
            % \rowcolor{blue!10}
            \rule{0pt}{10pt}{DDNM \cite{wang2022ddnm}} &{46.73} / {0.992} / {0.011}& 33.51 / 0.907 / {0.111}
            &{39.37} / {0.982} / {0.017} &31.64 / 0.885 / {0.116} & {28.49} / {26.19} / {0.090}\\
            
            \rowcolor{orange!8}
            \rule{0pt}{10pt}{DDNM\textbf{-EquS$^+$}} & {46.88} / {0.993} / {0.010} &{34.76} / {0.926} / {0.091} 
            &{39.85} / {0.983} / {0.016} &{32.44} / {0.901} / 0.126 &{31.01} / {26.08} / {0.086}\\
        \hline
    \end{tabular}
    \caption{Quantitative results of zero-shot DMIR methods on  \textbf{ImageNet}(\textit{top}) and \textbf{CelebA}(\textit{bottom}), including five typical IR tasks. {\colorbox{orange!8}{The enhanced versions}} show improved quantitative performance over their original counterparts.}
    %We mark {bold} for the best scores, {underline} for the second-best scores, and \protect {waveunderline} for the third-best scores, respectively.
    \label{tb:ndm}
\end{table*}
%-------------------------------------------------------------------------

\section{Experiments}

%-------------------------------------------------------------------------
\subsection{Experiment Details}

%-------------------------------------------------------------------------
\noindent\textbf{Experiment Settings.}
We evaluate EquS and EquS$^+$ across a range of typical IR tasks, including compressed sensing (CS), inpainting, super-resolution (SR), deblurring, and colorization. For validation, we use the ImageNet 1K \cite{deng2009imagenet} and CelebA 1K \cite{karras2018progressive} datasets, with images resized to 256×256. To ensure fair comparisons, we apply the same pre-trained diffusion models across all methods: \cite{dhariwal2021diffusion} for ImageNet and \cite{lugmayr2022repaint} for CelebA-HQ. All experiments were performed on NVIDIA GeForce RTX 4090. During implementation, EquS uses horizontal flipping to construct $\F$, while TAS employs a quadratic sampling schedule without altering the length of the sub-sequence $S$.
We append a suffix to the method name when incorporating our approach. Due to space constraints, additional details on the experiments can be found in the Supplementary.
%-------------------------------------------------------------------------

\noindent\textbf{Evaluation Metrics.}
We employ PSNR, SSIM \cite{wang2004image}, and LPIPS \cite{zhang2018unreasonable} as the main evaluation metrics for most IR tasks. For colorization specifically, we utilize the Consistency metric (referred to as \textit{Cons}) \cite{wang2022ddnm} along with FID \cite{fid}, since PSNR and SSIM do not adequately capture performance in this context. Generally, improved performance is indicated by higher PSNR and SSIM values, alongside lower values for LPIPS, FID, and \textit{Cons}.

%------------------------------------------------------------------------

\subsection{Evaluation on Different IR Tasks}

%-------------------------------------------------------------------------
We compare our method against SOTA zero-shot DMIR methods, including DPS \cite{chung2023dps}, DDRM \cite{kawar2022ddrm}, DDNM \cite{wang2022ddnm}, \(\Pi\)GDM \cite{song2023pseudoinverse}, CDDB \cite{cao2024deep}, and DeqIR \cite{cao2024deep}. EquS$^+$ is integrated into DDRM, CDDB, and DDNM. We evaluate under five typical IR tasks: block-based CS at a 0.25 compression ratio, text mask for inpainting, 4$\times$ bicubic downsampling for SR, Gaussian blur for deblurring, and average grayscale operator for colorization. $\A^\dagger \y$ are used as baselines.

Additional quantitative and qualitative results, including varying degrees of deconstruction and different IR tasks (\eg Walsh-Hadamard CS, anisotropic blur), are provided in the Supplementary.

%-------------------------------------------------------------------------

\noindent\textbf{The Effect of Our Method.} We show the quantitative results of zero-shot DMIR methods in Table \ref{tb:ndm}. {\colorbox{orange!8}{The enhanced versions}} show improved performance over their original counterparts. In Figure \ref{fig:overview}, we show that our method offers better quantitative performance and improved qualitative results.
For example, on the ImageNet for the Gaussian deblurring task, DDNM-EquS$^+$ achieved a significant performance boost over DDNM, with a notable 2.05 dB increase in PSNR and a 0.005 improvement in LPIPS; On the ImageNet for block-based CS task, CDDB-EquS$^+$ achieves a PSNR improvement of 3.45 dB and an LPIPS improvement of 0.126 compared to CDDB.
By utilizing dual-trajectory sampling and TAS, EquS$^+$ significantly enhances previous SOTA methods, demonstrating both the effectiveness (in quantitative performance improvement) and versatility (across various IR tasks) of our method.

%-------------------------------------------------------------------------
\begin{figure*}
\centering
\vspace{-5mm}
\includegraphics[width=1\linewidth]{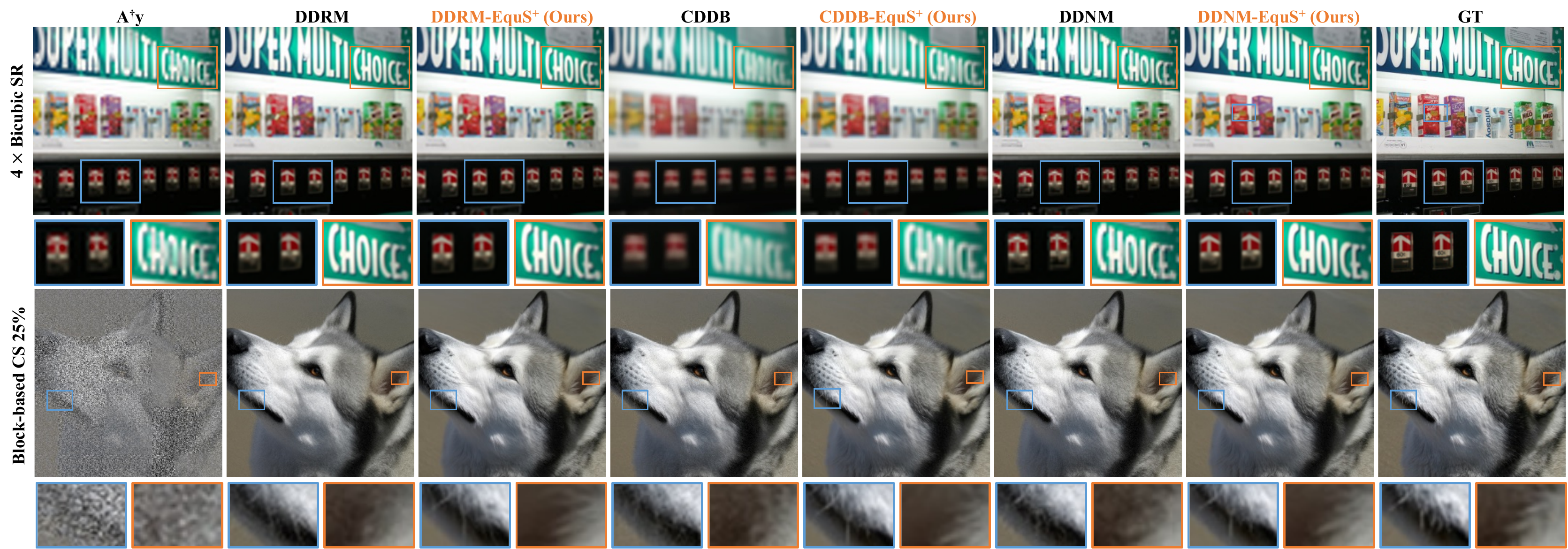}
\vspace{-2mm}
\caption{Qualitative results of zero-shot DMIR methods. A clearer text (\textit{top}) and the fur of the animals (\textit{bottom}) can be observed.}
\label{fig:qualitative}
\end{figure*}

%-------------------------------------------------------------------------
\noindent\textbf{Quantitative and Qualitative Results.} 
\textbf{CS.} CDDB-EquS$^+$ outperforms most other methods on benchmarks. Compared to the competitive IR method \({\Pi}\)GDM, CDDB-EquS$^+$ achieves an LPIPS improvement of up to 0.071 and a PSNR gain of up to 1.44 dB on ImageNet. 
For qualitative results, as shown in Figure \ref{fig:qualitative}, {\colorbox{orange!8}{the enhanced versions}} demonstrate enhanced visual quality with finely restored details of animal fur. These visual comparisons align closely with the quantitative findings, underscoring the effectiveness of our approach.
\textbf{Deblurring.} Notably, DDNM-EquS$^+$ outperforms most other methods on benchmarks.
\textbf{Inpainting.} CDDB-EquS$^+$ outperforms most other methods on benchmarks. For instance, compared to the competitive IR method DPS, CDDB-EquS$^+$ achieves an LPIPS improvement of up to 0.031 and a PSNR gain of up to 0.57 dB on ImageNet.
\textbf{SR.} DDNM-EquS$^+$ outperforms most other methods on CelebA. Although \({\Pi}\)GDM achieves superior performance on the ImageNet dataset, its reliance on gradient computation introduces additional computational overhead. For qualitative results, as shown in Figure \ref{fig:qualitative}, {\colorbox{orange!8}{the enhanced versions}} demonstrate better visual quality, capturing more legible text. Our method achieves substantial performance gains without additional computational cost.
\textbf{Colorization.} In Table \ref{tb:ndm}, the enhanced versions show competitive performance, which implies that there exists an inevitable trade-off between distortion (\textit{Cons}) and perception (FID, LPIPS) \cite{blau2018perception} in colorization.
In Figure \ref{fig:overview}, we further show that our method is adaptable to various IR applications, robust to different scales, and resilient to different noise levels. Due to space constraints, we provided more qualitative results in the Supplementary.

%-------------------------------------------------------------------------
% \begin{figure}[ht]
%     \centering
%     \vspace{-4mm}
%     \includegraphics[width=\linewidth]{pic/noisy_DDPG.pdf}
%     \vspace{-5mm}
%     \caption{Qualitative results of noisy IR tasks on CelebA-HQ.}
%     \vspace{-1mm}
%     \label{fig:noisy_DDPG}
% \end{figure}
\begin{table}
    \centering
    % \footnotesize
    \scriptsize
    \resizebox{\linewidth}{!}{
    \renewcommand{\arraystretch}{1.5}
    % \vspace{-3mm}
    \begin{tabular}{cc|ccc}
        \hline
           \multicolumn{2}{c}{\scriptsize\textbf{CelebA}}&\multicolumn{1}{c}{$\mathbf{A}^{\dagger}\mathbf{y}$} &\multicolumn{1}{c}{DDPG \cite{garber2024image}}&\multicolumn{1}{c}{\cellcolor{orange!8}{DDPG\textbf{-EquS}}}\\
           \hline
           Task & $\sigma_{\y}$ & PSNR↑/SSIM↑/LPIPS↓ & PSNR↑/SSIM↑/LPIPS↓ & \cellcolor{orange!8}{PSNR↑/SSIM↑/LPIPS↓} \\
        \hline
            \rule{0pt}{6pt}{\multirow{4}{*}{GD}} & 0 &36.58 / 0.934 / 0.124 &47.76 / 0.995 / 0.006
            &\cellcolor{orange!8}{\textbf{47.77 }/ \underline{0.995} / \underline{0.006}}\\
            & 0.01 &22.83 / 0.514 / 0.410 &33.06 / 0.903 / 0.113& \cellcolor{orange!8}{\textbf{33.14} / \textbf{0.904} / \textbf{0.112}}\\
            & 0.05 &19.74 / 0.306 / 0.552 &31.22 / 0.868 / 0.144&\cellcolor{orange!8}{ \textbf{31.32} / \textbf{0.870} / \textbf{0.143}}\\
            & 0.10 &17.79 / 0.201 / 0.623 &29.94 / 0.823 / 0.180&\cellcolor{orange!8}{\textbf{30.04} / \textbf{0.825} / \textbf{0.179}}\\
            \hline
            {\multirow{3}{*}{MD}}
            & 0.01 &21.77 / 0.486 / 0.461 &32.54 / 0.890 / 0.126&\cellcolor{orange!8}{\textbf{32.60} / \textbf{0.891} / \textbf{0.125}}\\
            & 0.05 &17.17 / 0.206 / 0.638 &29.01 / 0.817 / 0.177&\cellcolor{orange!8}{\textbf{29.17} / \textbf{0.821} / \textbf{0.175}}\\
            & 0.10 &18.85 / 0.686 / 0.305 &27.71 / 0.779 / 0.200
            &\cellcolor{orange!8}{\textbf{27.91} / \textbf{0.785} / \textbf{0.197}}\\
            \hline
            {\multirow{3}{*}{SR}} & 0 &29.77 / 0.862 / 0.233 &31.59 / 0.886 / 0.120
            &\cellcolor{orange!8}{\textbf{31.73} / \textbf{0.888} / \textbf{0.119}}\\
            & 0.01 &29.28 / 0.828 / 0.260 &31.81 / 0.882 / 0.154&\cellcolor{orange!8}{\textbf{31.91} / \textbf{0.883} / \underline{0.154}}\\
            & 0.05 & 24.83 / 0.560 / 0.469 &29.39 / 0.828 / 0.188&\cellcolor{orange!8}{\textbf{29.52} / \textbf{0.831} / \textbf{0.187}}\\
        \hline
        \end{tabular}}
%     \resizebox{\linewidth}{!}{
%     \renewcommand{\arraystretch}{1.5}
% \begin{tabular}{cc|ccc}
%         \hline
%            \multicolumn{2}{c}{\scriptsize\textbf{ImageNet}}&\multicolumn{1}{c}{$\mathbf{A}^{\dagger}\mathbf{y}$} &\multicolumn{1}{c}{DDPG}&\multicolumn{1}{c}{\cellcolor{orange!8}{DDPG\textbf{-EquS}}}\\
%            \hline
%            Task & \sigma_{\y} & PSNR↑/SSIM↑/LPIPS↓ & PSNR↑/SSIM↑/LPIPS↓ & \cellcolor{orange!8}{PSNR↑/SSIM↑/LPIPS↓} \\
%         \hline
%             \rule{0pt}{6pt}{\multirow{2}{*}{GD}} & 0 &32.72 / 0.874 / 0.163 &47.30 / 0.996 / 0.004
%             &\cellcolor{orange!8}{\textbf{47.31} / \underline{0.996} / \underline{0.004}}\\
%             & 0.05 &18.81 / 0.316 / 0.537 &28.77 / 0.827 / 0.194&\cellcolor{orange!8}{\textbf{28.79} / \underline{0.827} / \textbf{0.193}}\\
%             \hline
%             {\multirow{2}{*}{MD}} &0.05&16.81 / 0.204 / 0.622 &25.93 / 0.724 / 0.286&\cellcolor{orange!8}{ \textbf{25.94} / \textbf{0.725} / \textbf{0.285}}\\
%             & 0.10 &15.20 / 0.122 / 0.669 & 24.19 / 0.636 / 0.364
%             &\cellcolor{orange!8}{\textbf{24.20} / \textbf{0.637 }/ \textbf{0.363}} \\
%             \hline
%             {\multirow{2}{*}{SR}} & 0 &25.64 / 0.743 / 0.322 &27.40 / 0.801 / 0.237
%             &\cellcolor{orange!8}{ \textbf{27.41} / \underline{0.801} / \underline{0.237}}\\
%             & 0.05 &22.72 / 0.504 / 0.496 &25.55 / 0.716 / 0.342&\cellcolor{orange!8}{\textbf{25.56} / \textbf{0.717} / \textbf{0.343}}\\
%         \hline
%     \end{tabular}}
\vspace{-2mm}
    \caption{Quantitative results of noisy IR tasks on CelebA-HQ. $\sigma_{\y}$ denotes the noise level. ``GD'' represents Gaussian Deblurring, ``MD'' represents Motion Deblurring, and ``SR'' represents 4$\times$ Bicubic SR. We use \textbf{bold} to indicate improved scores and \underline{underline} to represent results that match previous methods.}
    \label{tb:DDPG}
\end{table}
%-------------------------------------------------------------------------
\setlength{\aboverulesep}{0.1pt} 
\setlength{\belowrulesep}{0.1pt}
\begin{table}[t]
    \centering
    % \footnotesize
    \scriptsize
    \resizebox{\linewidth}{!}{
    \renewcommand{\arraystretch}{1.5}
\begin{tabular}{cccc}
        \hline
           \multicolumn{1}{c}{\scriptsize\textbf{ImageNet}}&\multicolumn{1}{c}{$\mathbf{A}^{\dagger}\mathbf{y}$} &\multicolumn{1}{c}{DDNM$^+$ \cite{wang2022ddnm}}&\cellcolor{orange!8}{DDNM$^+$\textbf{-EquS}} \\
              Task & PSNR↑/SSIM↑/LPIPS↓ & PSNR↑/SSIM↑/LPIPS↓ & \cellcolor{orange!8}{PSNR↑/SSIM↑/LPIPS↓} \\
        \hline
            CS  & 14.22 / 0.154 / 0.680& 20.68 / 0.588 / 0.418 &\cellcolor{orange!8}{\textbf{21.22} / \textbf{0.604} / \textbf{0.412 }}\\
            SR & 14.33 / 0.149 / 0.749 & 22.66 / 0.604 / 0.420 &\cellcolor{orange!8}{\textbf{25.71} / \textbf{0.724} / \textbf{0.329}} \\
            % Colorization & 13.94 / 0.195 / 0.750& 17.39 / 0.404 / 0.574 &&\cellcolor{blue!5}{\textbf{ 21.31} / \textbf{0.716} / \textbf{0.395}}\\
           \cmidrule{1-4}
             & \textit{Cons}↓/FID↓/LPIPS↓ & \textit{Cons}↓/FID↓/LPIPS↓ & \cellcolor{orange!8}{\textit{Cons}↓/FID↓/LPIPS↓} \\
             \cmidrule{2-4}
           Colorization & 9376 / 239.4 / 0.750 & 3362 / 125.1 / 0.574 & \cellcolor{orange!8}{\textbf{1969} / \textbf{88.76 }/ \textbf{0.395}} \\
        \hline
\end{tabular}}
% \setlength{\aboverulesep}{0.1pt} 
% \setlength{\belowrulesep}{0.1pt}
%     \resizebox{\linewidth}{!}{
%     \renewcommand{\arraystretch}{1.5}
% \begin{tabular}{cccc}
%         \hline
%            \multicolumn{1}{c}{\scriptsize\textbf{CelebA}}&\multicolumn{1}{c}{$\mathbf{A}^{\dagger}\mathbf{y}$} &\multicolumn{1}{c}{DDNM$^+$}&\cellcolor{orange!8}{DDNM$^+$\textbf{+EquS}}\\
%               Task & PSNR↑/SSIM↑/LPIPS↓ & PSNR↑/SSIM↑/LPIPS↓ & \cellcolor{orange!8}{PSNR↑/SSIM↑/LPIPS↓} \\
%         \hline
%            CS & 13.87 / 0.127 / 0.716& 24.96 / 0.761 / 0.240 &\cellcolor{orange!8}{\textbf{25.50} / \textbf{0.772} / \textbf{0.239}}\\
%            SR & 14.76 / 0.138 / 0.776 & 25.78 / 0.744 / 0.225 &\cellcolor{orange!8}{\textbf{26.05} / \textbf{0.757} / 0.242} \\
%            Inpainting & 12.04 / 0.116 / 0.744 & 30.50 / 0.860 / 0.167 & \cellcolor{orange!8}{\textbf{30.57} / \textbf{0.862} / 0.170} \\
%            \cmidrule{1-4}
%              & \textit{Cons}↓/FID↓/LPIPS↓ & \textit{Cons}↓/FID↓/LPIPS↓ & \cellcolor{orange!8}{\textit{Cons}↓/FID↓/LPIPS↓} \\
%              \cmidrule{2-4}
%            Colorization & 9274 / 288.7 / 0.793 & 1451 / 45.94 / 0.251 & \cellcolor{orange!8}{\textbf{1439} / 46.47 / \textbf{0.242}} \\
%         \hline
% \end{tabular}}
\vspace{-2mm}
    \caption{Quantitative results of noisy IR tasks on ImageNet. Noise level is 0.2. We use \textbf{bold} to indicate improved scores. ``CS'' represents Walshhadamard CS 25\%, and ``SR'' represents 4$\times$ Average Pooling SR.}
    \label{tb:ddnm+}
\end{table}
%-------------------------------------------------------------------------

\begin{figure*}
    \centering
    \includegraphics[width=\linewidth]{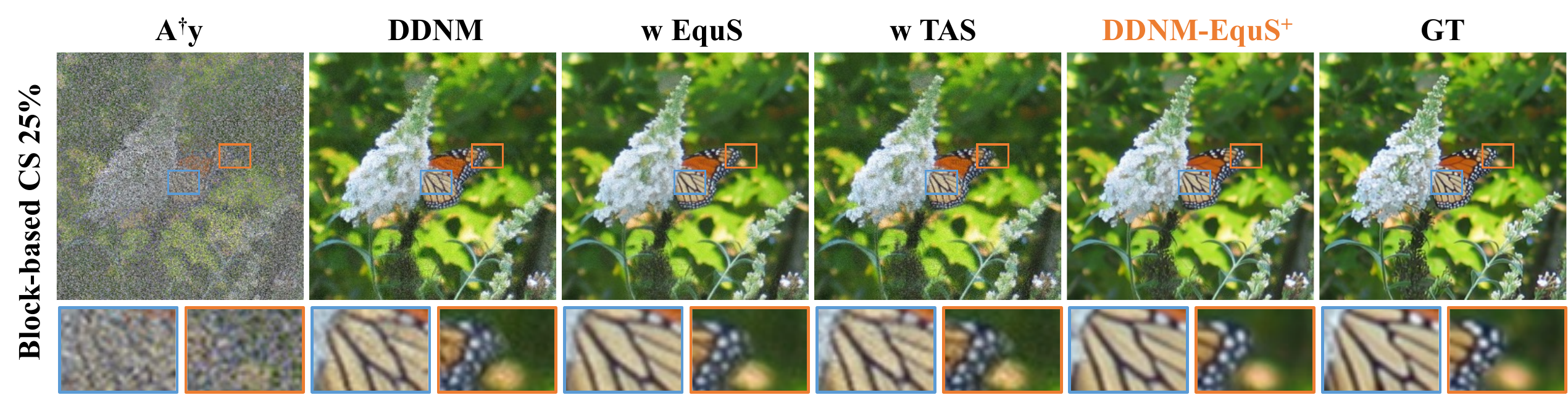}
    \caption{Qualitative results of ablation studies. The final version reveals clearer textures of the butterfly wings.}
    \label{fig:ablation}
\end{figure*}
%-------------------------------------------------------------------------
\subsection{Robustness in Noisy IR Tasks}

We also performed a comparison with DDPG \cite{garber2024image}, a SOTA zero-shot DMIR method for deblurring and SR, and validated the robustness of the proposed EquS under varying noise intensities. In Table \ref{tb:DDPG}, we show quantitative results on CelebA-HQ. {\colorbox{orange!8}{DDPG-EquS}} achieves better quantitative results than DDPG. These results indicate that \textit{EquS demonstrates stability across varying noise intensities}.

%-------------------------------------------------------------------------
\begin{figure}
    \centering
    \vspace{-3mm}
    \includegraphics[width=\linewidth]{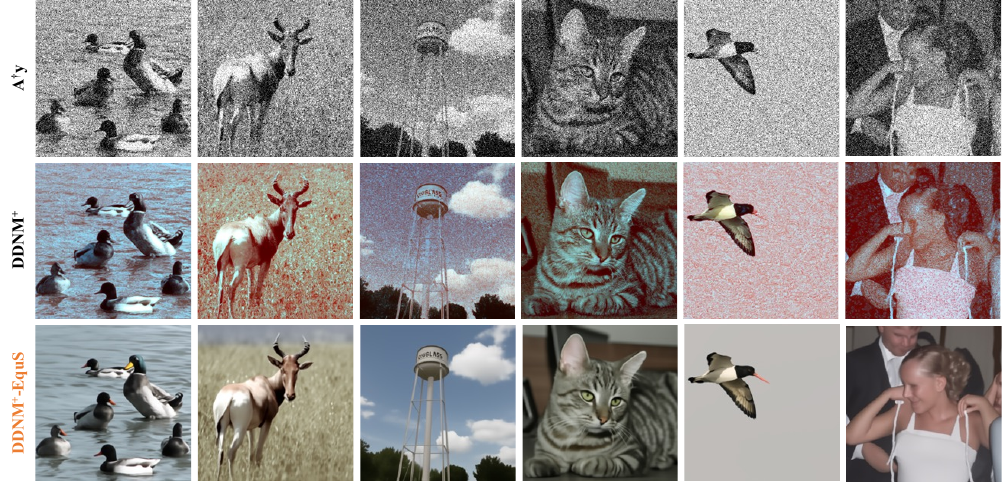}
    \vspace{-2mm}
    \caption{Qualitative results of colorization on Imagenet. Our method enables better visual quality.}
    \label{fig:DDNM+}
\end{figure}

Moreover, we validate EquS on DDNM$^+$ \cite{wang2022ddnm}, a method that can handle noisy IR tasks. We present the quantitative results on ImageNet in Table \ref{tb:ddnm+}. EquS significantly improves quantitative performance, such as achieving a 36.34 improvement in FID for colorization compared to DDNM$^+$. In Figure \ref{fig:DDNM+}, {\colorbox{orange!8}{DDNM$^+$-EquS}} achieves better visual quality.
Note that \textit{EquS can be seamlessly incorporated into previous methods without adding extra computational costs}.
% \subsection{Real World Applications}
% Our method can be applied in real-world settings.
% \begin{figure}[h]
%     \centering
%     \includegraphics[width=\linewidth]{pic/real-world.pdf}
%     \vspace{-4mm}
%     \caption{Real-world applications of our method.}
%     \vspace{-4mm}
%     \label{fig:real-world}
% \end{figure}
%-------------------------------------------------------------------------
\subsection{Test on other cases}

\noindent\textbf{Inpainting with Black Hole Mask.} To further demonstrate the robustness of the method, we experimented with two black hole masks in inpainting. The quantitative results are shown in Table \ref{tb:inpainting}. Our method is robust in these situations.
%We show qualitative results in the Supplementary due to space constraints.

\setlength{\aboverulesep}{0.1pt} 
\setlength{\belowrulesep}{0.1pt}
\begin{table}[t]
    \centering
    % \footnotesize
    \scriptsize
    \resizebox{\linewidth}{!}{
    \renewcommand{\arraystretch}{1.5}
\begin{tabular}{ccc}
        \hline
           \multicolumn{1}{c}{\scriptsize\textbf{CelebA}} &\multicolumn{1}{c}{DDNM \cite{wang2022ddnm}}&\cellcolor{orange!8}{DDNM\textbf{-EquS$^+$}} \\
              Inpainting & PSNR↑/SSIM↑/LPIPS↓ & \cellcolor{orange!8}{PSNR↑/SSIM↑/LPIPS↓} \\
        \hline
            Square & 35.29 / 0.977 / 0.018 &\cellcolor{orange!8}{\textbf{36.33} / \textbf{0.981} / \textbf{0.017}}\\
            Circle & 33.61 / 0.971 / 0.022 &\cellcolor{orange!8}{\textbf{34.61} / \textbf{0.975} / \textbf{0.021}} \\
        \hline
\end{tabular}}
    \caption{Quantitative results on ImageNet. We use \textbf{bold} to indicate improved scores. Our method is robust in these situations.}
    \label{tb:inpainting}
\end{table}

\subsection{Ablation Studies}

% In this paper, we propose EquS and TAS. To demonstrate their effectiveness,
\noindent\textbf{Proposed Methods.} To demonstrate the effect of our proposed EquS and TAS,
we performed ablation studies using DDNM \cite{wang2022ddnm} as a baseline.
The quantitative results shown in Table \ref{tb:ablation} highlight the benefits of the proposed EquS and TAS, resulting in consistent improvements in various IR tasks. We present qualitative results of ablation studies in Figure \ref{fig:ablation}, showing that EquS$^+$ captures finer texture details.
In summary, both EquS and EquS$^+$ have two notable advantages: \textbf{(1)} compatibility with previous methods and \textbf{(2)} no additional computational overhead.
Further ablation studies on more IR tasks are available in the Supplementary.

\begin{figure}
  \centering
  \begin{subfigure}{0.49\linewidth}
    \includegraphics[width=1\linewidth]{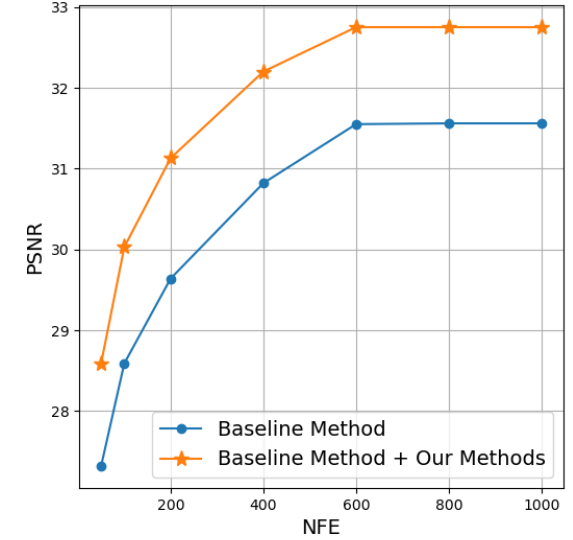}
    \caption{NFE vs. PSNR (\textbf{higher} is better).}
  \end{subfigure}
  \hfill
  \begin{subfigure}{0.49\linewidth}
    \includegraphics[width=1\linewidth]{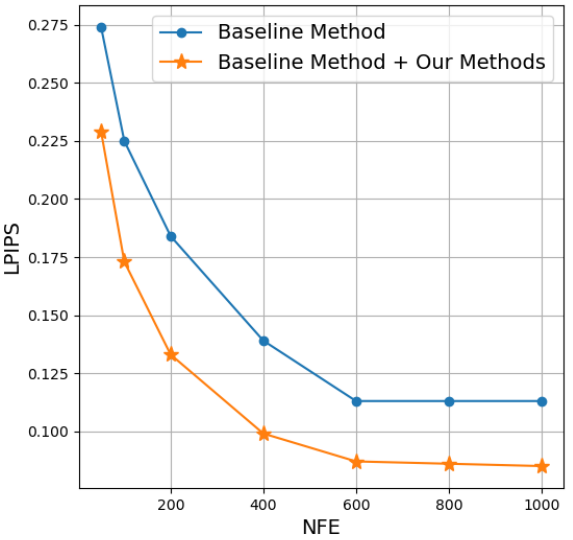}
    \caption{NFE vs. LPIPS (\textbf{lower} is better).}
  \end{subfigure}
  \vspace{-2mm}
  \caption{NFE vs. Evaluation metrics on block-based CS 25\%. Our method is not limited by specific NFE.}
  \vspace{-2mm}
  \label{fig:NFEvsEvaluation}
\end{figure}

\noindent\textbf{NFE vs. Evaluation Metrics.} In Figure \ref{fig:NFEvsEvaluation}, it can be seen that our method consistently improves the quantitative results under different NFE conditions. This indicates that our method is not limited by the number of sampling steps.

\begin{table}
\resizebox{\linewidth}{!}{
    \renewcommand{\arraystretch}{1.5}
\begin{tabular}{ccccc}
        \hline
        \multicolumn{2}{c}{\textbf{ImageNet}} &\multicolumn{1}{c}{Gaussian Deblurring} &\multicolumn{1}{c}{Block-based CS 25\%} &\multicolumn{1}{c}{Inpainting} \\
% &\multicolumn{1}{c}{Bicubic SR} 
           EquS & TAS &  PSNR↑/SSIM↑/LPIPS↓ & PSNR↑/SSIM↑/LPIPS↓ & PSNR↑/SSIM↑/LPIPS↓ \\
        \hline
             \textcolor[rgb]{1,0,0}{\ding{56}} &\textcolor[rgb]{1,0,0}{\ding{56}} &44.94 / 0.990 / 0.011 &28.59 / 0.795 / 0.225 & 34.93 / 0.968 / 0.025 \\
            \textcolor[rgb]{0,0.5,0}{\ding{52}} &\textcolor[rgb]{1,0,0}{\ding{56}} & 45.37 / 0.991 / 0.009 & 30.02 / 0.845 / 0.173 & 35.65 / 0.973 / 0.023 \\
            \textcolor[rgb]{1,0,0}{\ding{56}} & \textcolor[rgb]{0,0.5,0}{\ding{52}} & 46.31 / 0.992 / 0.007 & 30.32 / 0.844 / 0.153 & 35.59 / 0.972 / \textbf{0.021}\\
            \textcolor[rgb]{0,0.5,0}{\ding{52}} & \textcolor[rgb]{0,0.5,0}{\ding{52}} & \textbf{46.99} / \textbf{0.993} / \textbf{0.006 }& \textbf{31.87} / \textbf{0.887} / \textbf{0.103} & \textbf{36.13} / \textbf{0.976} / \textbf{0.021}  \\
        \hline
        % \multicolumn{2}{c}{Config2} & 27.91 / 0.807 / 0.251  & 30.34 / 0.845 / 0.152 & 35.60 / 0.972 / 0.022 \\
    \end{tabular}}
    \caption{Quantitative results of ablation studies. We use \textbf{bold} to indicate the best scores.}
    \label{tb:ablation}
\end{table}

%% file: sec/6_conclusion.tex
\section{Conclusion}
% In this paper, we have proposed a novel zero-shot diffusion model-based IR method, called EquS.
% Specifically, we incorporate an equivariant inverse mapping into the sampling process and impose bidirectional information via dual sampling trajectories.
% Furthermore, we introduce TAS to assign more weight to deterministic steps and incorporate more certainty in sampling. 
% With TAS, EquS$^+$ further improve the overall efficiency of the sampling chain and the image restoration performance.
% Comprehensive experiments validate the outstanding performance of EquS and EquS$^+$ across various IR tasks.
% Moreover, our method can be seamlessly integrated into previous methods and improve their performance at no additional costs.

In this paper, we propose a novel zero-shot diffusion model-based IR method, called EquS. Specifically, we incorporate an equivariant inverse mapping into the sampling process and impose equivariant information through dual sampling trajectories. Additionally, we introduce TAS to prioritize deterministic steps and enhance certainty in the sampling process. With TAS, EquS$^+$ further improves both the overall efficiency of the sampling chain and the IR performance. Comprehensive experiments demonstrate the outstanding performance of EquS and EquS$^+$ across a range of IR tasks. Moreover, our methods can be seamlessly integrated into existing methods, boosting their performance without incurring additional computational costs.

% \section*{Broader impacts}
% \label{Broader impacts}
% EquS offers advantages as a simple, training-free solution that can be readily integrated with existing methods. This zero-shot approach lowers the barrier to implementing high-quality image restoration across various practical scenarios. However, we acknowledge the potential risk that malicious actors could potentially exploit the image inpainting capabilities to alter important images or documents. While our method is fundamentally designed for restoration rather than generation, we strongly caution against such misuse and recommend implementing proper verification protocols when applying this technology to sensitive content.

%% file: main.bib
@String(CVPR= {IEEE Conf. Comput. Vis. Pattern Recog.})

@String(ICCV= {Int. Conf. Comput. Vis.})

@String(ECCV= {Eur. Conf. Comput. Vis.})

@String(TIP  = {IEEE Trans. Image Process.})

@String(TVCG  = {IEEE Trans. Vis. Comput. Graph.})

@String(ICLR = {Int. Conf. Learn. Represent.})

@String(AAAI = {AAAI})

@String(CVPR  = {CVPR})

@String(ICCV  = {ICCV})

@String(ECCV  = {ECCV})

@String(TIP   = {IEEE TIP})

@String(TVCG  = {IEEE TVCG})

@String(TCSVT = {IEEE TCSVT})

@String(ICLR  = {ICLR})

@article{wang2004image,
  title={Image quality assessment: from error visibility to structural similarity},
  author={Wang, Zhou and Bovik, Alan C and Sheikh, Hamid R and Simoncelli, Eero P},
  journal={IEEE transactions on image processing},
  volume={13},
  number={4},
  pages={600--612},
  year={2004},
  publisher={IEEE}
}

@inproceedings{chung2023solving,
  title={Solving 3d inverse problems using pre-trained 2d diffusion models},
  author={Chung, Hyungjin and Ryu, Dohoon and McCann, Michael T and Klasky, Marc L and Ye, Jong Chul},
  booktitle={Proceedings of the IEEE/CVF Conference on Computer Vision and Pattern Recognition},
  pages={22542--22551},
  year={2023}
}

@article{richardson1972bayesian,
  title={Bayesian-based iterative method of image restoration},
  author={Richardson, William Hadley},
  journal={JoSA},
  volume={62},
  number={1},
  pages={55--59},
  year={1972},
  publisher={Optica Publishing Group}
}

@inproceedings{terris2024equivariant,
  title={Equivariant plug-and-play image reconstruction},
  author={Terris, Matthieu and Moreau, Thomas and Pustelnik, Nelly and Tachella, Julian},
  booktitle={Proceedings of the IEEE/CVF Conference on Computer Vision and Pattern Recognition},
  pages={25255--25264},
  year={2024}
}

@article{banham1997digital,
  title={Digital image restoration},
  author={Banham, Mark R and Katsaggelos, Aggelos K},
  journal={IEEE signal processing magazine},
  volume={14},
  number={2},
  pages={24--41},
  year={1997},
  publisher={IEEE}
}

@inproceedings{cao2024deep,
  title={Deep Equilibrium Diffusion Restoration with Parallel Sampling},
  author={Cao, Jiezhang and Shi, Yue and Zhang, Kai and Zhang, Yulun and Timofte, Radu and Van Gool, Luc},
  booktitle={CVPR},
  pages={2824--2834},
  year={2024}
}

@inproceedings{whang2022deblurring,
  title={Deblurring via stochastic refinement},
  author={Whang, Jay and Delbracio, Mauricio and Talebi, Hossein and Saharia, Chitwan and Dimakis, Alexandros G and Milanfar, Peyman},
  booktitle={Proceedings of the IEEE/CVF Conference on Computer Vision and Pattern Recognition},
  pages={16293--16303},
  year={2022}
}

@article{alberti2021learning,
  title={Learning the optimal Tikhonov regularizer for inverse problems},
  author={Alberti, Giovanni S and De Vito, Ernesto and Lassas, Matti and Ratti, Luca and Santacesaria, Matteo},
  journal={Advances in Neural Information Processing Systems},
  volume={34},
  pages={25205--25216},
  year={2021}
}

@inproceedings{liu2024residual,
  title={Residual denoising diffusion models},
  author={Liu, Jiawei and Wang, Qiang and Fan, Huijie and Wang, Yinong and Tang, Yandong and Qu, Liangqiong},
  booktitle={Proceedings of the IEEE/CVF Conference on Computer Vision and Pattern Recognition},
  pages={2773--2783},
  year={2024}
}

@article{lehtinen2018noise2noise,
  title={Noise2Noise: Learning image restoration without clean data},
  author={Lehtinen, Jaakko and Munkberg, Jacob and Hasselgren, Jon and Laine, Samuli and Karras, Tero and Aittala, Miika and Aila, Timo},
  journal={arXiv preprint arXiv:1803.04189},
  year={2018}
}

@inproceedings{batson2019noise2self,
  title={Noise2self: Blind denoising by self-supervision},
  author={Batson, Joshua and Royer, Loic},
  booktitle={International Conference on Machine Learning},
  pages={524--533},
  year={2019},
  organization={PMLR}
}

@inproceedings{krull2019noise2void,
  title={Noise2void-learning denoising from single noisy images},
  author={Krull, Alexander and Buchholz, Tim-Oliver and Jug, Florian},
  booktitle={Proceedings of the IEEE/CVF conference on computer vision and pattern recognition},
  pages={2129--2137},
  year={2019}
}

@inproceedings{moran2020noisier2noise,
  title={Noisier2noise: Learning to denoise from unpaired noisy data},
  author={Moran, Nick and Schmidt, Dan and Zhong, Yu and Coady, Patrick},
  booktitle={Proceedings of the IEEE/CVF Conference on Computer Vision and Pattern Recognition},
  pages={12064--12072},
  year={2020}
}

@article{xu2020noisy,
  title={Noisy-as-clean: Learning self-supervised denoising from corrupted image},
  author={Xu, Jun and Huang, Yuan and Cheng, Ming-Ming and Liu, Li and Zhu, Fan and Xu, Zhou and Shao, Ling},
  journal={IEEE Transactions on Image Processing},
  volume={29},
  pages={9316--9329},
  year={2020},
  publisher={IEEE}
}

@inproceedings{wang2023noise2info,
  title={Noise2info: Noisy image to information of noise for self-supervised image denoising},
  author={Wang, Jiachuan and Di, Shimin and Chen, Lei and Ng, Charles Wang Wai},
  booktitle={Proceedings of the IEEE/CVF International Conference on Computer Vision},
  pages={16034--16043},
  year={2023}
}

@inproceedings{mansour2023zero,
  title={Zero-shot noise2noise: Efficient image denoising without any data},
  author={Mansour, Youssef and Heckel, Reinhard},
  booktitle={Proceedings of the IEEE/CVF Conference on Computer Vision and Pattern Recognition},
  pages={14018--14027},
  year={2023}
}

@inproceedings{sohl2015deep,
  title={Deep unsupervised learning using nonequilibrium thermodynamics},
  author={Sohl-Dickstein, Jascha and Weiss, Eric and Maheswaranathan, Niru and Ganguli, Surya},
  booktitle={International conference on machine learning},
  pages={2256--2265},
  year={2015},
  organization={PMLR}
}

@article{song2019generative,
  title={Generative modeling by estimating gradients of the data distribution},
  author={Song, Yang and Ermon, Stefano},
  journal={Advances in neural information processing systems},
  volume={32},
  year={2019}
}

@article{croitoru2023diffusion,
  title={Diffusion models in vision: A survey},
  author={Croitoru, Florinel-Alin and Hondru, Vlad and Ionescu, Radu Tudor and Shah, Mubarak},
  journal={IEEE Transactions on Pattern Analysis and Machine Intelligence},
  year={2023},
  publisher={IEEE}
}

@article{li2022srdiff,
  title={Srdiff: Single image super-resolution with diffusion probabilistic models},
  author={Li, Haoying and Yang, Yifan and Chang, Meng and Chen, Shiqi and Feng, Huajun and Xu, Zhihai and Li, Qi and Chen, Yueting},
  journal={Neurocomputing},
  volume={479},
  pages={47--59},
  year={2022},
  publisher={Elsevier}
}

@inproceedings{gao2023implicit,
  title={Implicit diffusion models for continuous super-resolution},
  author={Gao, Sicheng and Liu, Xuhui and Zeng, Bohan and Xu, Sheng and Li, Yanjing and Luo, Xiaoyan and Liu, Jianzhuang and Zhen, Xiantong and Zhang, Baochang},
  booktitle={Proceedings of the IEEE/CVF conference on computer vision and pattern recognition},
  pages={10021--10030},
  year={2023}
}

@article{chen2024hierarchical,
  title={Hierarchical integration diffusion model for realistic image deblurring},
  author={Chen, Zheng and Zhang, Yulun and Liu, Ding and Gu, Jinjin and Kong, Linghe and Yuan, Xin and others},
  journal={Advances in neural information processing systems},
  volume={36},
  year={2024}
}

@inproceedings{ren2023multiscale,
  title={Multiscale structure guided diffusion for image deblurring},
  author={Ren, Mengwei and Delbracio, Mauricio and Talebi, Hossein and Gerig, Guido and Milanfar, Peyman},
  booktitle={Proceedings of the IEEE/CVF International Conference on Computer Vision},
  pages={10721--10733},
  year={2023}
}

@inproceedings{corneanu2024latentpaint,
  title={Latentpaint: Image inpainting in latent space with diffusion models},
  author={Corneanu, Ciprian and Gadde, Raghudeep and Martinez, Aleix M},
  booktitle={Proceedings of the IEEE/CVF Winter Conference on Applications of Computer Vision},
  pages={4334--4343},
  year={2024}
}

@inproceedings{carrillo2023diffusart,
  title={Diffusart: Enhancing line art colorization with conditional diffusion models},
  author={Carrillo, Hernan and Cl{\'e}ment, Micha{\"e}l and Bugeau, Aur{\'e}lie and Simo-Serra, Edgar},
  booktitle={Proceedings of the IEEE/CVF Conference on Computer Vision and Pattern Recognition},
  pages={3486--3490},
  year={2023}
}

@inproceedings{rombach2022high,
  title={High-resolution image synthesis with latent diffusion models},
  author={Rombach, Robin and Blattmann, Andreas and Lorenz, Dominik and Esser, Patrick and Ommer, Bj{\"o}rn},
  booktitle={Proceedings of the IEEE/CVF conference on computer vision and pattern recognition},
  pages={10684--10695},
  year={2022}
}

@article{dhariwal2021diffusion,
  title={Diffusion models beat gans on image synthesis},
  author={Dhariwal, Prafulla and Nichol, Alexander},
  journal={Advances in neural information processing systems},
  volume={34},
  pages={8780--8794},
  year={2021}
}

@article{weng2024cad,
  title={L-cad: Language-based colorization with any-level descriptions using diffusion priors},
  author={Weng, Shuchen and Zhang, Peixuan and Li, Yu and Li, Si and Shi, Boxin and others},
  journal={Advances in Neural Information Processing Systems},
  volume={36},
  year={2024}
}

@article{yang2024lossy,
  title={Lossy image compression with conditional diffusion models},
  author={Yang, Ruihan and Mandt, Stephan},
  journal={Advances in Neural Information Processing Systems},
  volume={36},
  year={2024}
}

@article{kim2021noise2score,
  title={Noise2score: tweedie’s approach to self-supervised image denoising without clean images},
  author={Kim, Kwanyoung and Ye, Jong Chul},
  journal={Advances in Neural Information Processing Systems},
  volume={34},
  pages={864--874},
  year={2021}
}

@article{hyvarinen2005estimation,
  title={Estimation of non-normalized statistical models by score matching.},
  author={Hyv{\"a}rinen, Aapo and Dayan, Peter},
  journal={Journal of Machine Learning Research},
  volume={6},
  number={4},
  year={2005}
}

@inproceedings{ho2020ddpm,
  title={Denoising diffusion probabilistic models},
  author={Ho, Jonathan and Jain, Ajay and Abbeel, Pieter},
  booktitle={Advances in Neural Information Processing Systems},
  year={2020}
}

@inproceedings{song2020ddim,
  title={Denoising diffusion implicit models},
  author={Song, Jiaming and Meng, Chenlin and Ermon, Stefano},
  booktitle={International conference on machine learning},
  year={2021}
}

@inproceedings{chung2023dps,
  title={Diffusion Posterior Sampling for General Noisy Inverse Problems},
  author={Chung, Hyungjin and Kim, Jeongsol and Mccann, Michael T and Klasky, Marc L and Ye, Jong Chul},
  booktitle={ICLR},
  year={2023}
}

@inproceedings{kawar2022ddrm,
  title={Denoising diffusion restoration models},
  author={Kawar, Bahjat and Elad, Michael and Ermon, Stefano and Song, Jiaming},
  booktitle={Advances in Neural Information Processing Systems},
  year={2022}
}

@inproceedings{wang2022ddnm,
  title={Zero-shot image restoration using denoising diffusion null-space model},
  author={Wang, Yinhuai and Yu, Jiwen and Zhang, Jian},
  booktitle={International conference on machine learning},
  year={2023}
}

@inproceedings{song2020score,
  title={Score-based generative modeling through stochastic differential equations},
  author={Song, Yang and Sohl-Dickstein, Jascha and Kingma, Diederik P and Kumar, Abhishek and Ermon, Stefano and Poole, Ben},
  booktitle={International conference on machine learning},
  year={2021}
}

@inproceedings{zhu2023diffpir,
  title={Denoising Diffusion Models for Plug-and-Play Image Restoration},
  author={Zhu, Yuanzhi and Zhang, Kai and Liang, Jingyun and Cao, Jiezhang and Wen, Bihan and Timofte, Radu and Van Gool, Luc},
  booktitle={Proceedings of the IEEE/CVF Conference on Computer Vision and Pattern RecognitionW},
  year={2023}
}

@article{saharia2022sr3,
  title={Image super-resolution via iterative refinement},
  author={Saharia, Chitwan and Ho, Jonathan and Chan, William and Salimans, Tim and Fleet, David J and Norouzi, Mohammad},
  journal={IEEE Transactions on Pattern Analysis and Machine Intelligence},
  year={2022},
}

@inproceedings{saharia2022palette,
  title={Palette: Image-to-image diffusion models},
  author={Saharia, Chitwan and Chan, William and Chang, Huiwen and Lee, Chris and Ho, Jonathan and Salimans, Tim and Fleet, David and Norouzi, Mohammad},
  booktitle={ACM SIGGRAPH},
  year={2022}
}

@inproceedings{xia2023diffir,
  title={Diffir: Efficient diffusion model for image restoration},
  author={Xia, Bin and Zhang, Yulun and Wang, Shiyin and Wang, Yitong and Wu, Xinglong and Tian, Yapeng and Yang, Wenming and Van Gool, Luc},
  booktitle={ICCV},
  year={2023}
}

@inproceedings{lugmayr2022repaint,
  title={Repaint: Inpainting using denoising diffusion probabilistic models},
  author={Lugmayr, Andreas and Danelljan, Martin and Romero, Andres and Yu, Fisher and Timofte, Radu and Van Gool, Luc},
  booktitle={Proceedings of the IEEE/CVF Conference on Computer Vision and Pattern Recognition},
  year={2022}
}

@inproceedings{choi2021ilvr,
  title={ILVR: Conditioning Method for Denoising Diffusion Probabilistic Models},
  author={Choi, Jooyoung and Kim, Sungwon and Jeong, Yonghyun and Gwon, Youngjune and Yoon, Sungroh},
  booktitle={ICCV},
  year={2021}
}

@inproceedings{menon2020pulse,
  title={Pulse: Self-supervised photo upsampling via latent space exploration of generative models},
  author={Menon, Sachit and Damian, Alexandru and Hu, Shijia and Ravi, Nikhil and Rudin, Cynthia},
  booktitle={Proceedings of the IEEE/CVF Conference on Computer Vision and Pattern Recognition},
  year={2020}
}

@inproceedings{zhang2018unreasonable,
  title={The unreasonable effectiveness of deep features as a perceptual metric},
  author={Zhang, Richard and Isola, Phillip and Efros, Alexei A and Shechtman, Eli and Wang, Oliver},
  booktitle={Proceedings of the IEEE conference on computer vision and pattern recognition},
  pages={586--595},
  year={2018}
}

@inproceedings{rombach2022ldm,
  title={High-resolution image synthesis with latent diffusion models},
  author={Rombach, Robin and Blattmann, Andreas and Lorenz, Dominik and Esser, Patrick and Ommer, Bj{\"o}rn},
  booktitle={Proceedings of the IEEE/CVF Conference on Computer Vision and Pattern Recognition},
  year={2022}
}

@article{lin2023diffbir,
  title={DiffBIR: Towards Blind Image Restoration with Generative Diffusion Prior},
  author={Lin, Xinqi and He, Jingwen and Chen, Ziyan and Lyu, Zhaoyang and Fei, Ben and Dai, Bo and Ouyang, Wanli and Qiao, Yu and Dong, Chao},
  journal={arXiv preprint arXiv:2308.15070},
  year={2023}
}

@inproceedings{zhang2021bsrgan,
    title={Designing a Practical Degradation Model for Deep Blind Image Super-Resolution},
    author={Zhang, Kai and Liang, Jingyun and Van Gool, Luc and Timofte, Radu},
    booktitle={ICCV},
    year={2021}
}

@article{dong2015srcnn,
  title={Image super-resolution using deep convolutional networks},
  author={Dong, Chao and Loy, Chen Change and He, Kaiming and Tang, Xiaoou},
  journal={IEEE Transactions on Pattern Analysis and Machine Intelligence},
  year={2015}
}

@inproceedings{dong201arcnn,
  title={Compression artifacts reduction by a deep convolutional network},
  author={Dong, Chao and Deng, Yubin and Loy, Chen Change and Tang, Xiaoou},
  booktitle={ICCV},
  year={2015}
}

@inproceedings{VDSR,
  title={Accurate image super-resolution using very deep convolutional networks},
  author={Kim, Jiwon and Lee, Jung Kwon and Lee, Kyoung Mu},
  booktitle={Proceedings of the IEEE/CVF Conference on Computer Vision and Pattern Recognition},
  year={2016}
}

@article{plug-denoiser,
  title={Plug-and-play image restoration with deep denoiser prior},
  author={Zhang, Kai and Li, Yawei and Zuo, Wangmeng and Zhang, Lei and Van Gool, Luc and Timofte, Radu},
  journal={IEEE Transactions on Pattern Analysis and Machine Intelligence},
  year={2021},
}

@inproceedings{WGAN-GP,
  title={Improved training of wasserstein gans},
  author={Gulrajani, Ishaan and Ahmed, Faruk and Arjovsky, Martin and Dumoulin, Vincent and Courville, Aaron},
  booktitle={Advances in Neural Information Processing Systems},
  year={2017}
}

@inproceedings{ESRGAN,
  title={Esrgan: Enhanced super-resolution generative adversarial networks},
  author={Wang, Xintao and Yu, Ke and Wu, Shixiang and Gu, Jinjin and Liu, Yihao and Dong, Chao and Qiao, Yu and Change Loy, Chen},
  booktitle={ECCVW},
  year={2018}
}

@inproceedings{Real-ESRGAN,
  title={Real-ESRGAN: Training real-world blind super-resolution with pure synthetic data},
  author={Wang, Xintao and Xie, Liangbin and Dong, Chao and Shan, Ying},
  booktitle={ICCVW},
  year={2021}
}

@inproceedings{inpainting-GAN,
  title={Context encoders: Feature learning by inpainting},
  author={Pathak, Deepak and Krahenbuhl, Philipp and Donahue, Jeff and Darrell, Trevor and Efros, Alexei A},
  booktitle={Proceedings of the IEEE/CVF Conference on Computer Vision and Pattern Recognition},
  year={2016}
}

@article{goodfellow2014generative,
  title={Generative adversarial nets},
  author={Goodfellow, Ian and Pouget-Abadie, Jean and Mirza, Mehdi and Xu, Bing and Warde-Farley, David and Ozair, Sherjil and Courville, Aaron and Bengio, Yoshua},
  journal={Advances in neural information processing systems},
  volume={27},
  year={2014}
}

@article{pan2021exploiting,
  title={Exploiting deep generative prior for versatile image restoration and manipulation},
  author={Pan, Xingang and Zhan, Xiaohang and Dai, Bo and Lin, Dahua and Loy, Chen Change and Luo, Ping},
  journal={IEEE Transactions on Pattern Analysis and Machine Intelligence},
  year={2021},
  publisher={IEEE}
}

@inproceedings{inpainting3,
  title={Image inpainting with learnable bidirectional attention maps},
  author={Xie, Chaohao and Liu, Shaohui and Li, Chao and Cheng, Ming-Ming and Zuo, Wangmeng and Liu, Xiao and Wen, Shilei and Ding, Errui},
  booktitle={ICCV},
  year={2019}
}

@inproceedings{deepfillv2,
  title={Free-form image inpainting with gated convolution},
  author={Yu, Jiahui and Lin, Zhe and Yang, Jimei and Shen, Xiaohui and Lu, Xin and Huang, Thomas S},
  booktitle={ICCV},
  year={2019}
}

@inproceedings{restormer,
  title={Restormer: Efficient transformer for high-resolution image restoration},
  author={Zamir, Syed Waqas and Arora, Aditya and Khan, Salman and Hayat, Munawar and Khan, Fahad Shahbaz and Yang, Ming-Hsuan},
  booktitle={Proceedings of the IEEE/CVF Conference on Computer Vision and Pattern Recognition},
  year={2022}
}

@inproceedings{MAT,
  title={Mat: Mask-aware transformer for large hole image inpainting},
  author={Li, Wenbo and Lin, Zhe and Zhou, Kun and Qi, Lu and Wang, Yi and Jia, Jiaya},
  booktitle={Proceedings of the IEEE/CVF Conference on Computer Vision and Pattern Recognition},
  year={2022}
}

@inproceedings{NAFNet,
  title={Simple baselines for image restoration},
  author={Chen, Liangyu and Chu, Xiaojie and Zhang, Xiangyu and Sun, Jian},
  booktitle={ECCV},
  year={2022},
}

@inproceedings{restor12,
  title={Focnet: A fractional optimal control network for image denoising},
  author={Jia, Xixi and Liu, Sanyang and Feng, Xiangchu and Zhang, Lei},
  booktitle={Proceedings of the IEEE/CVF Conference on Computer Vision and Pattern Recognition},
  year={2019}
}

@article{restor14,
  title={A pseudo-blind convolutional neural network for the reduction of compression artifacts},
  author={Kim, Yoonsik and Soh, Jae Woong and Park, Jaewoo and Ahn, Byeongyong and Lee, Hyun-Seung and Moon, Young-Su and Cho, Nam Ik},
  journal={TCSVT},
  year={2019}
}

@article{restor15,
  title={A model-driven deep unfolding method for jpeg artifacts removal},
  author={Fu, Xueyang and Wang, Menglu and Cao, Xiangyong and Ding, Xinghao and Zha, Zheng-Jun},
  journal={TNNLS},
  year={2021}
}

@article{AOTGAN,
  title={Aggregated contextual transformations for high-resolution image inpainting},
  author={Zeng, Yanhong and Fu, Jianlong and Chao, Hongyang and Guo, Baining},
  journal={TVCG},
  year={2022}
}

@article{liang2022vrt,
  title={Vrt: A video restoration transformer},
  author={Liang, Jingyun and Cao, Jiezhang and Fan, Yuchen and Zhang, Kai and Ranjan, Rakesh and Li, Yawei and Timofte, Radu and Van Gool, Luc},
  journal={TIP},
  year={2024}
}

@article{cao2021vsrt,
  title={Video super-resolution transformer},
  author={Cao, Jiezhang and Li, Yawei and Zhang, Kai and Van Gool, Luc},
  journal={arXiv preprint arXiv:2106.06847},
  year={2021}
}

@InProceedings{cao2023ciaosr,
    author    = {Cao, Jiezhang and Wang, Qin and Xian, Yongqin and Li, Yawei and Ni, Bingbing and Pi, Zhiming and Zhang, Kai and Zhang, Yulun and Timofte, Radu and Van Gool, Luc},
    title     = {CiaoSR: Continuous Implicit Attention-in-Attention Network for Arbitrary-Scale Image Super-Resolution},
    booktitle = {Proceedings of the IEEE/CVF Conference on Computer Vision and Pattern Recognition},
    year      = {2023}
}

@inproceedings{cao2022datsr,
  title={Reference-based image super-resolution with deformable attention transformer},
  author={Cao, Jiezhang and Liang, Jingyun and Zhang, Kai and Li, Yawei and Zhang, Yulun and Wang, Wenguan and Gool, Luc Van},
  booktitle={ECCV},
  year={2022}
}

@inproceedings{cao2022davsr,
  title={Towards interpretable video super-resolution via alternating optimization},
  author={Cao, Jiezhang and Liang, Jingyun and Zhang, Kai and Wang, Wenguan and Wang, Qin and Zhang, Yulun and Tang, Hao and Van Gool, Luc},
  booktitle={ECCV},
  year={2022},
}

@inproceedings{deng2009imagenet,
  title={Imagenet: A large-scale hierarchical image database},
  author={Deng, Jia and Dong, Wei and Socher, Richard and Li, Li-Jia and Li, Kai and Fei-Fei, Li},
  booktitle={Proceedings of the IEEE/CVF Conference on Computer Vision and Pattern Recognition},
  year={2009},
}

@inproceedings{wang2023stablesr,
    author = {Wang, Jianyi and Yue, Zongsheng and Zhou, Shangchen and Chan, Kelvin CK and Loy, Chen Change},
    title = {Exploiting Diffusion Prior for Real-World Image Super-Resolution},
    booktitle = {arXiv preprint arXiv:2305.07015},
    year = {2023}
}

@article{yue2022difface,
  title={DifFace: Blind Face Restoration with Diffused Error Contraction},
  author={Yue, Zongsheng and Loy, Chen Change},
  journal={arXiv preprint arXiv:2212.06512},
  year={2022}
}

@inproceedings{cao2018lccgan,
  title={Adversarial learning with local coordinate coding},
  author={Cao, Jiezhang and Guo, Yong and Wu, Qingyao and Shen, Chunhua and Huang, Junzhou and Tan, Mingkui},
  booktitle={International Conference on Machine Learning},
  pages={707--715},
  year={2018},
  organization={PMLR}
}

@article{cao2020improving,
  title={Improving generative adversarial networks with local coordinate coding},
  author={Cao, Jiezhang and Guo, Yong and Wu, Qingyao and Shen, Chunhua and Huang, Junzhou and Tan, Mingkui},
  journal={IEEE Transactions on Pattern Analysis and Machine Intelligence},
  year={2020}
}

@article{kawar2021snips,
  title={Snips: Solving noisy inverse problems stochastically},
  author={Kawar, Bahjat and Vaksman, Gregory and Elad, Michael},
  journal={Advances in Neural Information Processing Systems},
  year={2021}
}

@inproceedings{dou2024diffusion,
  title={Diffusion posterior sampling for linear inverse problem solving: A filtering perspective},
  author={Dou, Zehao and Song, Yang},
  booktitle={The Twelfth International Conference on Learning Representations},
  year={2024}
}

@inproceedings{dgp,
  author = {Pan, Xingang and Zhan, Xiaohang and Dai, Bo and Lin, Dahua and Loy, Chen Change and Luo, Ping},
  title = {Exploiting Deep Generative Prior for Versatile Image Restoration and Manipulation},
  booktitle = {European Conference on Computer Vision (ECCV)},
  year = {2020}
}

@inproceedings{liu2023i2sb,
  title={I2SB: image-to-image Schr{\"o}dinger bridge},
  author={Liu, Guan-Horng and Vahdat, Arash and Huang, De-An and Theodorou, Evangelos A and Nie, Weili and Anandkumar, Anima},
  booktitle={Proceedings of the 40th International Conference on Machine Learning},
  pages={22042--22062},
  year={2023}
}

@article{chung2024deep,
  title={Deep diffusion image prior for efficient ood adaptation in 3d inverse problems},
  author={Chung, Hyungjin and Ye, Jong Chul},
  journal={arXiv preprint arXiv:2407.10641},
  year={2024}
}

@article{alccalar2024zero,
  title={Zero-shot adaptation for approximate posterior sampling of diffusion models in inverse problems},
  author={Al{\c{c}}alar, Ya{\c{s}}ar Utku and Ak{\c{c}}akaya, Mehmet},
  journal={arXiv preprint arXiv:2407.11288},
  year={2024}
}

@inproceedings{blau2018perception,
  title={The perception-distortion tradeoff},
  author={Blau, Yochai and Michaeli, Tomer},
  booktitle={Proceedings of the IEEE Conference on Computer Vision and Pattern Recognition},
  pages={6228--6237},
  year={2018}
}

@inproceedings{imagenet,
  title={{ImageNet: A large-scale hierarchical image database}},
  author={Deng, Jia and Dong, Wei and Socher, Richard and Li, Li-Jia and Li, Kai and Fei-Fei, Li},
  booktitle={2009 IEEE Conference on Computer Vision and Pattern Recognition},
  pages={248--255},
  year={2009}
}

@inproceedings{fid,
 author = {Heusel, Martin and Ramsauer, Hubert and Unterthiner, Thomas and Nessler, Bernhard and Hochreiter, Sepp},
 booktitle = {Advances in Neural Information Processing Systems},
 title = {GANs Trained by a Two Time-Scale Update Rule Converge to a Local Nash Equilibrium},
 volume = {30},
 year = {2017}
}

@string{eccv={ECCV}}

@string{ICLR={International conference on machine learning}}

@string{iccv={ICCV}}

@string{siggraph={SIGGRAPH}}

@article{vincent2011connection,
  title={A connection between score matching and denoising autoencoders},
  author={Vincent, Pascal},
  journal={Neural computation},
  volume={23},
  number={7},
  pages={1661--1674},
  year={2011},
  publisher={MIT Press}
}

@inproceedings{karras2018progressive,
    title={Progressive Growing of {GAN}s for Improved Quality, Stability, and Variation},
    author={Tero Karras and Timo Aila and Samuli Laine and Jaakko Lehtinen},
    booktitle={International conference on machine learning},
    year={2018}
}

@inproceedings{wu2024id,
  title={ID-Blau: Image Deblurring by Implicit Diffusion-based reBLurring AUgmentation},
  author={Wu, Jia-Hao and Tsai, Fu-Jen and Peng, Yan-Tsung and Tsai, Chung-Chi and Lin, Chia-Wen and Lin, Yen-Yu},
  booktitle={Proceedings of the IEEE/CVF Conference on Computer Vision and Pattern Recognition},
  pages={25847--25856},
  year={2024}
}

@inproceedings{garber2024image,
  title={Image restoration by denoising diffusion models with iteratively preconditioned guidance},
  author={Garber, Tomer and Tirer, Tom},
  booktitle={Proceedings of the IEEE/CVF Conference on Computer Vision and Pattern Recognition},
  pages={25245--25254},
  year={2024}
}

@inproceedings{wang2023unlimited,
  title={Unlimited-size diffusion restoration},
  author={Wang, Yinhuai and Yu, Jiwen and Yu, Runyi and Zhang, Jian},
  booktitle={Proceedings of the IEEE/CVF Conference on Computer Vision and Pattern Recognition},
  pages={1160--1167},
  year={2023}
}

@inproceedings{chen2022robust,
  title={Robust equivariant imaging: a fully unsupervised framework for learning to image from noisy and partial measurements},
  author={Chen, Dongdong and Tachella, Juli{\'a}n and Davies, Mike E},
  booktitle={Proceedings of the IEEE/CVF Conference on Computer Vision and Pattern Recognition},
  pages={5647--5656},
  year={2022}
}

@article{chen2023imaging,
  title={Imaging With Equivariant Deep Learning: From unrolled network design to fully unsupervised learning},
  author={Chen, Dongdong and Davies, Mike and Ehrhardt, Matthias J and Sch{\"o}nlieb, Carola-Bibiane and Sherry, Ferdia and Tachella, Juli{\'a}n},
  journal={IEEE Signal Processing Magazine},
  volume={40},
  number={1},
  pages={134--147},
  year={2023},
  publisher={IEEE}
}

@inproceedings{chen2021equivariant,
  title={Equivariant imaging: Learning beyond the range space},
  author={Chen, Dongdong and Tachella, Juli{\'a}n and Davies, Mike E},
  booktitle={Proceedings of the IEEE/CVF International Conference on Computer Vision},
  pages={4379--4388},
  year={2021}
}

@inproceedings{wang2023gan,
  title={Gan prior based null-space learning for consistent super-resolution},
  author={Wang, Yinhuai and Hu, Yujie and Yu, Jiwen and Zhang, Jian},
  booktitle={Proceedings of the AAAI Conference on Artificial Intelligence},
  volume={37},
  number={3},
  pages={2724--2732},
  year={2023}
}

@article{schwab2019deep,
  title={Deep null space learning for inverse problems: convergence analysis and rates},
  author={Schwab, Johannes and Antholzer, Stephan and Haltmeier, Markus},
  journal={Inverse Problems},
  volume={35},
  number={2},
  year={2019},
  publisher={IOP Publishing}
}

@inproceedings{gandikota2024text,
  title={Text-guided Explorable Image Super-resolution},
  author={Gandikota, Kanchana Vaishnavi and Chandramouli, Paramanand},
  booktitle={Proceedings of the IEEE/CVF Conference on Computer Vision and Pattern Recognition},
  pages={25900--25911},
  year={2024}
}

@article{chung2024direct,
  title={Direct diffusion bridge using data consistency for inverse problems},
  author={Chung, Hyungjin and Kim, Jeongsol and Ye, Jong Chul},
  journal={Advances in Neural Information Processing Systems},
  volume={36},
  year={2024}
}

@inproceedings{song2023pseudoinverse,
  title={Pseudoinverse-guided diffusion models for inverse problems},
  author={Song, Jiaming and Vahdat, Arash and Mardani, Morteza and Kautz, Jan},
  booktitle={ICLR},
  year={2023}
}

@inproceedings{ye2024learning,
  title={Learning Diffusion Texture Priors for Image Restoration},
  author={Ye, Tian and Chen, Sixiang and Chai, Wenhao and Xing, Zhaohu and Qin, Jing and Lin, Ge and Zhu, Lei},
  booktitle={Proceedings of the IEEE/CVF Conference on Computer Vision and Pattern Recognition},
  pages={2524--2534},
  year={2024}
}

@article{lee2024diffusion,
  title={Diffusion Prior-Based Amortized Variational Inference for Noisy Inverse Problems},
  author={Lee, Sojin and Park, Dogyun and Kong, Inho and Kim, Hyunwoo J},
  journal={arXiv preprint arXiv:2407.16125},
  year={2024}
}

@inproceedings{liu2024cdformer,
  title={CDFormer: When Degradation Prediction Embraces Diffusion Model for Blind Image Super-Resolution},
  author={Liu, Qingguo and Zhuang, Chenyi and Gao, Pan and Qin, Jie},
  booktitle={Proceedings of the IEEE/CVF Conference on Computer Vision and Pattern Recognition},
  pages={7455--7464},
  year={2024}
}

@inproceedings{li2024rethinking,
  title={Rethinking diffusion model for multi-contrast mri super-resolution},
  author={Li, Guangyuan and Rao, Chen and Mo, Juncheng and Zhang, Zhanjie and Xing, Wei and Zhao, Lei},
  booktitle={Proceedings of the IEEE/CVF Conference on Computer Vision and Pattern Recognition},
  pages={11365--11374},
  year={2024}
}

@inproceedings{zhou2020dudornet,
  title={DuDoRNet: learning a dual-domain recurrent network for fast MRI reconstruction with deep T1 prior},
  author={Zhou, Bo and Zhou, S Kevin},
  booktitle={Proceedings of the IEEE/CVF conference on computer vision and pattern recognition},
  pages={4273--4282},
  year={2020}
}

@inproceedings{dong2024building,
  title={Building Bridges across Spatial and Temporal Resolutions: Reference-Based Super-Resolution via Change Priors and Conditional Diffusion Model},
  author={Dong, Runmin and Yuan, Shuai and Luo, Bin and Chen, Mengxuan and Zhang, Jinxiao and Zhang, Lixian and Li, Weijia and Zheng, Juepeng and Fu, Haohuan},
  booktitle={Proceedings of the IEEE/CVF Conference on Computer Vision and Pattern Recognition},
  pages={27684--27694},
  year={2024}
}

@inproceedings{zhang2024diffusion,
  title={Diffusion-based Blind Text Image Super-Resolution},
  author={Zhang, Yuzhe and Zhang, Jiawei and Li, Hao and Wang, Zhouxia and Hou, Luwei and Zou, Dongqing and Bian, Liheng},
  booktitle={Proceedings of the IEEE/CVF Conference on Computer Vision and Pattern Recognition},
  pages={25827--25836},
  year={2024}
}

@inproceedings{lin2019dudonet,
  title={Dudonet: Dual domain network for ct metal artifact reduction},
  author={Lin, Wei-An and Liao, Haofu and Peng, Cheng and Sun, Xiaohang and Zhang, Jingdan and Luo, Jiebo and Chellappa, Rama and Zhou, Shaohua Kevin},
  booktitle={Proceedings of the IEEE/CVF Conference on Computer Vision and Pattern Recognition},
  pages={10512--10521},
  year={2019}
}

@article{meng2021sdedit,
  title={Sdedit: Guided image synthesis and editing with stochastic differential equations},
  author={Meng, Chenlin and He, Yutong and Song, Yang and Song, Jiaming and Wu, Jiajun and Zhu, Jun-Yan and Ermon, Stefano},
  journal={arXiv preprint arXiv:2108.01073},
  year={2021}
}
